\documentclass[letterpaper, 10 pt, journal, twoside]{IEEEtran}
\IEEEoverridecommandlockouts
\usepackage[utf8]{inputenc}
\usepackage[english]{babel}
\usepackage[T1]{fontenc}
\usepackage{amssymb,amsfonts}
\usepackage[cmex10]{amsmath}
\usepackage{dsfont}
\usepackage{algorithm}
\usepackage{algpseudocode}
\usepackage{multirow}
\usepackage{multicol}
\usepackage{graphicx}
\usepackage{textcomp}
\usepackage{xcolor}
\usepackage[binary-units=true]{siunitx}
\usepackage{tikz}
\usepackage[caption=false,font=footnotesize]{subfig}
\usepackage{hyperref}
\usepackage[capitalize,]{cleveref}
\crefformat{equation}{(#2#1#3)}
\crefrangeformat{equation}{(#3#1#4) to~(#5#2#6)}
\crefmultiformat{equation}{(#2#1#3)}%
{ and~(#2#1#3)}{, (#2#1#3)}{ and~(#2#1#3)}

\usepackage{array}
\usepackage{tabulary}
\usepackage{booktabs}
\newcolumntype{L}{>{$}l<{$}}
\newcolumntype{R}{>{$}r<{$}}
\newcolumntype{C}{>{$}c<{$}}

\def\BibTeX{{\rm B\kern-.05em{\sc i\kern-.025em b}\kern-.08em
    T\kern-.1667em\lower.7ex\hbox{E}\kern-.125emX}}
\newcommand {\real}{\mathbb{R}}

\newcommand\norm[1]{\left\lVert#1\right\rVert}
\newcommand{\disable}[1]{}
\newcommand{\Vmax}{\ensuremath{\text{v}\textsubscript{max}}}

\newcommand{\Amax}{\ensuremath{\text{a}\textsubscript{max}}}
\newcommand{\Kgain}{\ensuremath{\text{k}\textsubscript{gain}}}

\newcommand{\Dmin}{\ensuremath{\text{d}\textsubscript{min}}}
\newcommand{\matr}[1]{\mathbf{#1}}

\newcommand{\boundellipse}[3]% center, xdim, ydim
{(#1) ellipse (#2 and #3)
}

\newcommand{\reviewi}[1]{#1} %was brown
\newcommand{\reviewii}[1]{#1} %was blue

\DeclareMathOperator*{\optmin}{min}

\usepackage{scalerel}
\usetikzlibrary{svg.path}
\definecolor{orcidlogocol}{HTML}{A6CE39}
\tikzset{
orcidlogo/.pic={
  \fill[orcidlogocol] svg{M256,128c0,70.7-57.3,128-128,128C57.3,256,0,198.7,0,128C0,57.3,57.3,0,128,0C198.7,0,256,57.3,256,128z};
  \fill[white] svg{M86.3,186.2H70.9V79.1h15.4v48.4V186.2z}
               svg{M108.9,79.1h41.6c39.6,0,57,28.3,57,53.6c0,27.5-21.5,53.6-56.8,53.6h-41.8V79.1z M124.3,172.4h24.5c34.9,0,42.9-26.5,42.9-39.7c0-21.5-13.7-39.7-43.7-39.7h-23.7V172.4z}
               svg{M88.7,56.8c0,5.5-4.5,10.1-10.1,10.1c-5.6,0-10.1-4.6-10.1-10.1c0-5.6,4.5-10.1,10.1-10.1C84.2,46.7,88.7,51.3,88.7,56.8z};
}
}
\newcommand\orcidicon[1]{\href{https://orcid.org/#1}{\mbox{\scalerel*{
\begin{tikzpicture}[yscale=-1,transform shape]
\pic{orcidlogo};
\end{tikzpicture}
}{|}}}}
\begin{document}
\bstctlcite{etals}

\title{Multi-vehicle Dynamic Water Surface Monitoring}

\author{František Nekovář$^{\orcidicon{0000-0002-1975-078X}}$, \and Jan Faigl$^{\orcidicon{0000-0002-6193-0792}}$, \and Martin Saska$^{\orcidicon{0000-0001-7106-3816}}$
\thanks{
   Manuscript received: February 23, 2023; Revised June 17, 2023; Accepted July 21, 2023.}%
\thanks{
This paper was recommended for publication by
Editor Pauline Pounds upon evaluation of the Associate Editor and Reviewers’
comments.
   This work was supported by the Czech Science Foundation (GAČR) under research projects No. 22-05762S and No. 22-24425S, the CTU grant No. SGS23/177/OHK3/3T/13, and the Technology Innovation Institute - Sole Proprietorship LLC, UAE, under the Research Project Contract No. TII/ARRC/2055/2021.}
\thanks{Authors are with the Czech Technical University, Faculty of Electrical Engineering, Technicka 2, 166 27, Prague, Czech Republic, email: {\tt\small \{nekovfra|faiglj|saskam1\}@fel.cvut.cz}.}
\thanks{
  Digital Object Identifier (DOI): https://doi.org/10.1109/LRA.2023.3304533}
\thanks{© 2023 IEEE. Personal use of this material is permitted. Permission from IEEE must be
obtained for all other uses, in any current or future media, including
reprinting/republishing this material for advertising or promotional purposes, creating new
collective works, for resale or redistribution to servers or lists, or reuse of any copyrighted
component of this work in other works.}
}

\markboth{IEEE Robotics and Automation Letters. Preprint Version. Accepted July, 2023}
{Nekovář \MakeLowercase{\textit{et al.}}: Multi-vehicle Dynamic Water Surface Monitoring} 

\maketitle

\begin{abstract}
   Repeated exploration of a water surface to detect objects of interest and their subsequent monitoring is important in search-and-rescue or ocean clean-up operations.
Since the location of any detected object is dynamic, we propose to address the combined surface exploration and monitoring of the detected objects by modeling spatio-temporal reward states and coordinating a team of vehicles to collect the rewards.
The model characterizes the dynamics of the water surface and enables the planner to predict future system states.
The state reward value relevant to the particular water surface cell increases over time and is nullified by being in a sensor range of a vehicle.
Thus, the proposed multi-vehicle planning approach is to minimize the collective value of the dynamic model reward states.
The purpose is to address vehicles' motion constraints by using model predictive control on receding horizon and fully exploiting the utilized vehicles' motion capabilities.
Based on the evaluation results, the approach indicates improvement in a solution to the kinematic orienteering problem and the team orienteering problem in the monitoring task compared to the existing solutions.
The proposed approach has been experimentally verified, supporting its feasibility in real-world monitoring tasks.

\end{abstract}


% - section -------------------------------------------------------------------
\section{Introduction}
\IEEEPARstart{I}{n} this letter, we present a novel formulation of the model-based multi-vehicle planning, denoted \emph{Incremental Motion Planning with Dynamic Reward} (IMP-DR), to address continual exploration and monitoring of water surface with objects of interest using dynamic spatio-temporal reward model.
The studied problem belongs to a class of robotic scenarios in which repeated monitoring for state (or location) changes is needed upon encountering an object of interest during the initial exploration.
Besides, repeated monitoring can identify new objects that become detectable in time.

The motivational scenario is to employ a fleet of \emph{Unmanned Aerial Vehicles}~(UAVs) in top-down visual monitoring of large water surfaces surrounding a central ship, where the water surfaces that are not static due to tidal and weather conditions.
\reviewii{The problem is most closely related to the informative motion planning to maximize information gathering along the planned trajectory~\cite{hollinger14ijrr}, the kinematic \emph{Orienteering Problem} (OP) \cite{Meyer2022} to find vehicle's constrained reward-collecting trajectory and persistent monitoring with limited sensing range~\cite{smith2012persistent}.
Hence, we model the information gained from exploration and monitoring as dynamic reward states similar to the discretized field in~\cite{smith2012persistent}, where the states' position might change.}

The underlying spatio-temporal reward model dynamics influence the expected information gain, and we thus formulate the combined information-collecting task as maximizing the collected reward on a receding horizon with minimizing the weighted sum of the reward states.
While the value of any reward state steadily increases in time, it is nullified by its presence in the vehicle's sensor range.
Besides, the information gain is constrained by limited sensor range, and movement planning needs to consider vehicle motion constraints.

\begin{figure}[tb]\centering
   \includegraphics[width=0.95\columnwidth]{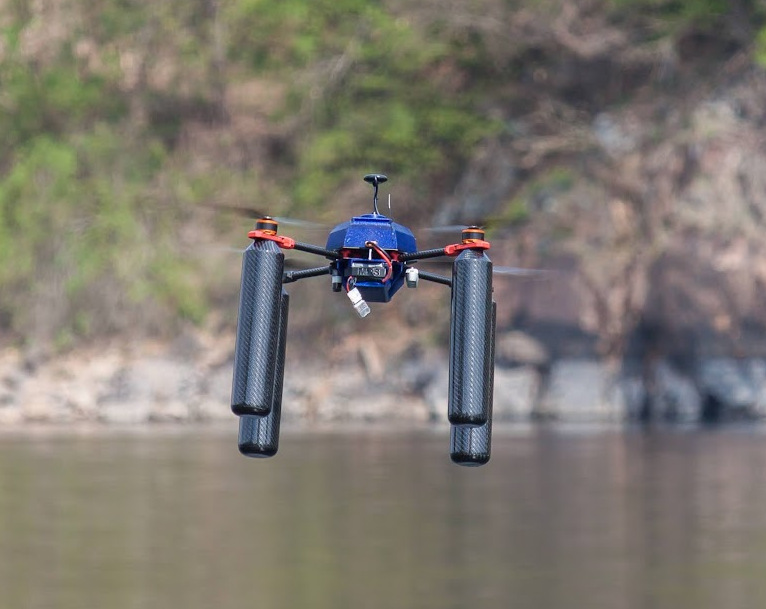}
   \vspace{-0.5em}
   \caption{Experimental UAV performing a water surface monitoring mission.\label{fig:uav_over_water}}
\end{figure}

\begin{figure}[tb]
   \includegraphics[width=\columnwidth]{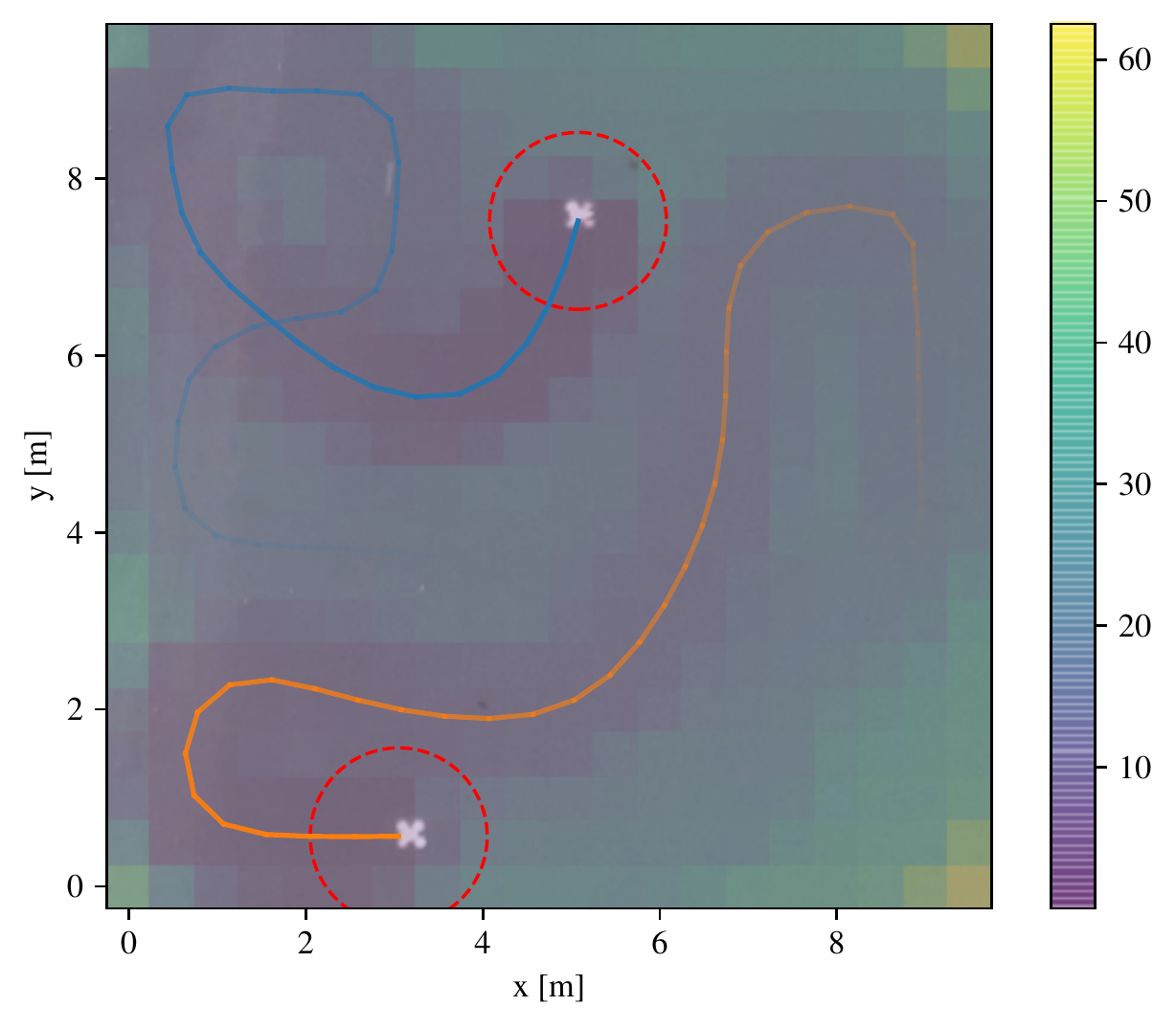}
   \vspace{-2.5em}
   \caption{Reward value overlay during monitoring experiment using two UAVs.
   The reward collection range is shown as red circles around the vehicles.
   \label{fig:plan_example}}
\end{figure}

We propose to address the studied problem using the \emph{Model Predictive Control} (MPC) approach that allows us to consider dynamical models of the vehicles, reward states, and positions of objects being monitored.
The dynamical models enable predicting future environment and vehicle states, with computational complexity exponentially increasing with prediction horizon length.
Thus, information-gathering trajectories are computed on a limited control horizon, and the iterative closed-loop operation scheme of the MPC planner with the system model is used for the continual exploration and monitoring using multiple vehicles.
The approach has been experimentally verified using a real vehicle shown in \cref{fig:uav_over_water}.
A~top-down snapshot from the deployment is depicted in Fig.~\ref{fig:plan_example} with overlaid reward states' values spaced on a symmetric grid.

The main contributions are considered as follows.
\begin{itemize}
   \item Novel IMP-DR problem formulation to coordinate a team of vehicles to search a priory unknown (water) surface and monitor detected dynamic objects continually.
   \item IMP-DR solution using MPC-based planning on receding horizon with a dynamic environment and vehicle models, predicting future states of the dynamic environment.
   \item Evaluation of the proposed solution and comparison with the most related (to the best of the authors' knowledge) approach to the \emph{Team Orienteering Problem} (TOP)~\cite{chao1996team} adapted for the receding horizon planning with dynamic model and solved by the state-of-the-art meta-heuristic~\cite{pedersen2021grasp}.
     The proposed approach is also compared to the state-of-the-art kinematic OP methods~\cite{Meyer2022} for a single vehicle showing improved results on evaluation scenarios.
   \item Experimental deployment of the proposed approach in real-world flight in a water surface monitoring scenario.
\end{itemize}

The remainder of the letter is organized as follows.
Related work is overviewed in the following section.
The addressed IMP-DR is formally introduced in \cref{sec:problem}.
The sensor and reward models are described in \cref{sec:models} together with the vehicle model and its motion constraints.
The proposed MPC-based solution is presented in \cref{sec:mpc}. 
Evaluation results and experimental deployment are reported in~\cref{sec:results}.
Concluding remarks are summarized in~\cref{sec:conclusion}.

% - section -------------------------------------------------------------------
\section{Related Work}\label{sec:related}

Environment monitoring can be formulated as a problem of creating a quantifiable phenomena model using sampled data from the environment~\cite{adaptive_sampling} with the regression about the phenomena state using regression models such as \emph{Gaussian Processes} (GPs)~\cite{tresp2000mixtures} and sensor model, such as in~\cite{tigris}.
Nevertheless, the studied continual monitoring task has to address vehicle motion constraints and information dynamics.
Therefore, path planning is a part of the navigation toward the locations where the studied phenomena can be measured, such as the signal strength~\cite{adaptive_informative}.
Thus, informative path planning~\cite{adaptive_selection} is to determine the most informative path improving the phenomena model with respect to motion constraints of information collecting vehicles~\cite{hollinger14ijrr,ghaffari2019sampling}.
However, these approaches are suitable for modeling spatial phenomena in environment monitoring tasks, but they do not address the underlying temporal dynamics that we call information dynamics and its interaction with the vehicles.

\reviewii{
Persistent multi-vehicle monitoring of changing environments with limited vehicle sensing range and pre-determined closed vehicles' paths is presented in~\cite{smith2012persistent}.
Continuous dynamic scalar field monitoring with value estimation based on individual filtered measurements is proposed in~\cite{lan2014variational}, where the authors employ Pontryagin's minimum principle for planning locally optimal, single-vehicle sensing trajectories minimizing the estimation uncertainty.
However, single-order integrator vehicle dynamics is considered with unconstrained input.
}%

In our addressed task, we explicitly model the environment information dynamics \reviewii{similarly to the accumulation function~\cite{smith2012persistent} in combination with sensor and vehicle dynamics} as a set of \emph{Discrete Algebraic Equations} (DAEs) used for moving-horizon prediction and control.
The limited prediction horizon allows planning to visit only some locations to collect measurements.
Hence, we need to select the most informative locations reachable, and the task can be formulated as a~generalization of the \emph{Orienteering Problem} (OP)~\cite{vansteenwegen_orienteering_2011}.

Although not directly addressing the information dynamics, existing OP formulations can be utilized in receding horizon reward collection planning schemes.
Multi-vehicle planning for search-and-rescue operations with grid-sampled rewards determined by satellite imaging of the area of interest is formulated in~\cite{pedersen2021grasp}.
However, exact visits of the sampled grid positions are required, constraining the vehicle's movement and neglecting its dynamics.
Non-zero sensor range and Dubins vehicle dynamics are addressed in~\cite{penivcka2019data}.
The authors of~\cite{Meyer2022} address the OP with time-optimal multi-rotor vehicle kinematic trajectory generation.
The OP with multiple Dubins vehicles si studied in~\cite{zahradka2018route}, providing background work on existing solutions to the planning problem with a limited travel budget.

Furthermore, we consider discretized environment representations inspiring, such as in exploring flooded areas~\cite{garg2022directed}, where the collection of the information reward is addressed by the uniform sampling of grid cells with travel budget-constrained vehicle-driving policy. 
\reviewii{The Correlated OP~\cite{yu2014correlated} formulates cyclic informative tours with the information gain correlated between neighboring nodes of the graph-based environment representation to estimate a scalar field.
However, the sensor range is limited to the individual nodes, and the approaches do not account for the vehicle dynamics.}

The dynamic position of the target objects is addressed in~\cite{khosravi2015maximum} by receding horizon planning and dynamic clustering to maximize the reward collected from moving targets with uncertain dynamics.
In~\cite{khosravi2016cooperative}, the approach is generalized to address obstacles in cooperative planning that can also be based on the Fisher information matrix model, for example, used in multi-UAV target tracking~\cite{koohifar2017receding}.

\reviewi{In~\cite{durgadoo2021strategies}, the dynamics of water surface debris show additional effects to the local oscillations to be taken into account, such as water currents, sail effect, and locally most dominant Stokes drift~\cite{van2018stokes}.
Therefore, we model the target movement as a combination of local oscillations and time-dependent drift.}

The existing receding horizon formulations address targets' and vehicles' position dynamics but not the reward values' dynamics for continual monitoring.
Therefore, we generalize the existing approaches to account for all the dynamics in the studied problem of combined water surface monitoring and tracking of the detected object of interest.

% - section -------------------------------------------------------------------
\section{Problem Statement}\label{sec:problem}

The \emph{Incremental Motion Planning with Dynamic Reward} (IMP-DR) problem is formulated as a multi-vehicle scenario in a two-dimensional environment represented by $n_p$ sampled locations or predicted positions of objects of interest, further referred to as targets, each encoded by a position vector $\matr{p}_i$.
Each $\matr{p}_i$ represents coordinates $\matr{p}_i = \left[p_{i,x},\, p_{i,y}\right]$, and targets form a discrete finite set $\mathcal{P} = \{ \matr{p}_i \in \real^2 : 1 \leq i \leq n_p \}$.
The number of targets $n_p$ can evolve during the mission as new targets might be added by the object detection.
\reviewi{Targets' coordinates can change with predictable dynamics, a model that is assumed to be known.
However, the model parameterization can change during the monitoring based on the visual observations.}
Thus, the target set is a function of time $\mathcal{P}(t)$.

The presence of $m$ vehicles is assumed at $\matr{q}_i = \left[q_{i,x},\,q_{i,y}\right]$ forming the set $\mathcal{Q} = \{ \matr{q}_i \in \real^2 : 1 \leq i \leq m  \}$.
Since the vehicles move in time, we denote $\mathcal{Q}(t)$.
Besides, the vehicles are subject to dynamical constraints given by the maximum velocity magnitude $\Vmax$ and acceleration in the axes $\Amax$.
Mutual vehicle collision avoidance and safety are addressed by the constraint on the minimum allowed distance between the vehicles $\Dmin$ at any time event $t$, such that $\norm{\matr{q}_i(t) - \matr{q}_j(t)} \geq \Dmin, \quad \forall \matr{q}_i(t), \matr{q}_j(t) \in \mathcal{Q}(t), i \neq j$, and $i,j\in\{1,\ldots, m\}$.

We seek to minimize the time elapsed between visits to the targets in $\mathcal{P}_t$ on a planning horizon $T_h$ while penalizing maximal value.
It is addressed by defining the dynamical reward state, analogous to the information in the informative planning approaches.
The reward states $r_i(t) \in \real^{+}_{0} : 0 \leq i \leq n_p$ are modeled for each $\matr{p}_i\in\mathcal{P}(t)$.
In the presence of any of the vehicles near the reward state position $\matr{p}_i$, its reward value $r_i$ is nullified.
Squared reward state value is used in optimization problems to penalize high values and improve solver performance.
The function $f_r(\dot{r},r,\mathcal{P},\mathcal{Q})$ describes implicit reward dynamics with relation to the targets and positions of the vehicles.
The variables of $f_r$ are functions of time; the notation is omitted for clarity.
We formulate the IMP-DR planning problem step as follows.
\begin{equation}\label{eq:impdr}
  \begin{aligned}
    \optmin_{\mathcal{Q}} & = \sum_{i=1}^{n_p} \int_{t_0}^{t_0 + T_h} r_i^2 dt \\
    \text{s.t.   } \\
     & 0 = f_r(\dot{r},r,\mathcal{P},\mathcal{Q}).
  \end{aligned}
\end{equation}

We use the IMP-DR to perform monitoring tasks by iterative solution of~\cref{eq:impdr}, incrementing the initial time $t_0$ with the sampling period $T_s \leq T_h$ on each iteration and utilizing the current reward state feedback.

% - section -------------------------------------------------------------------
\section{Dynamic Models}\label{sec:models}

The sensor model, reward model, and second-order multi-rotor UAV model used in the proposed monitoring solution are presented in this section.

% - subsection ----------------------------------------------------------------
\subsection{Sensor Model}
\begin{figure}[!t]\centering
   \subfloat[Butterworth function\label{fig:butter}]{\includegraphics[width=0.5\columnwidth]{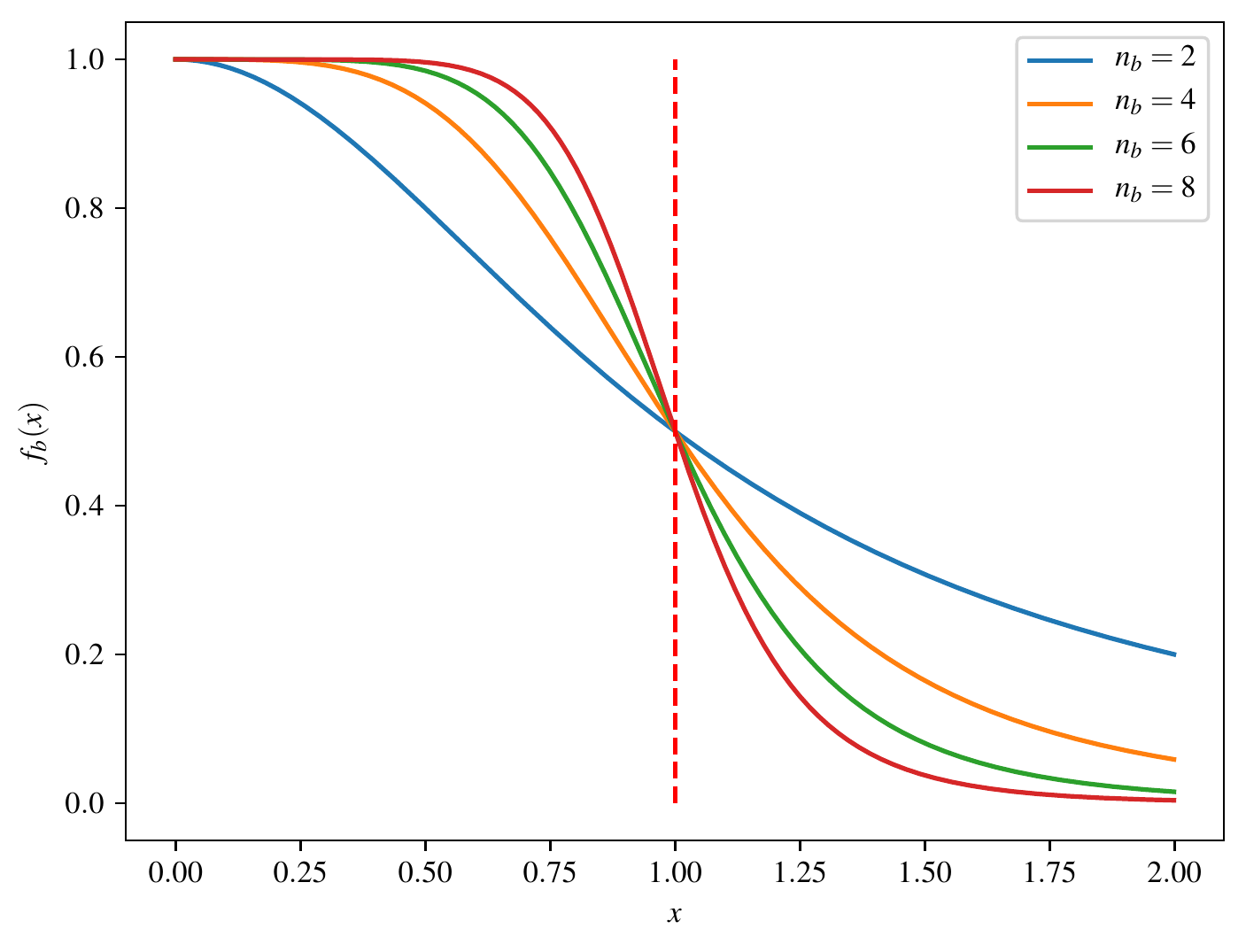}}
   \subfloat[Sensor range model\label{fig:visibility}]{\includegraphics[width=0.5\columnwidth]{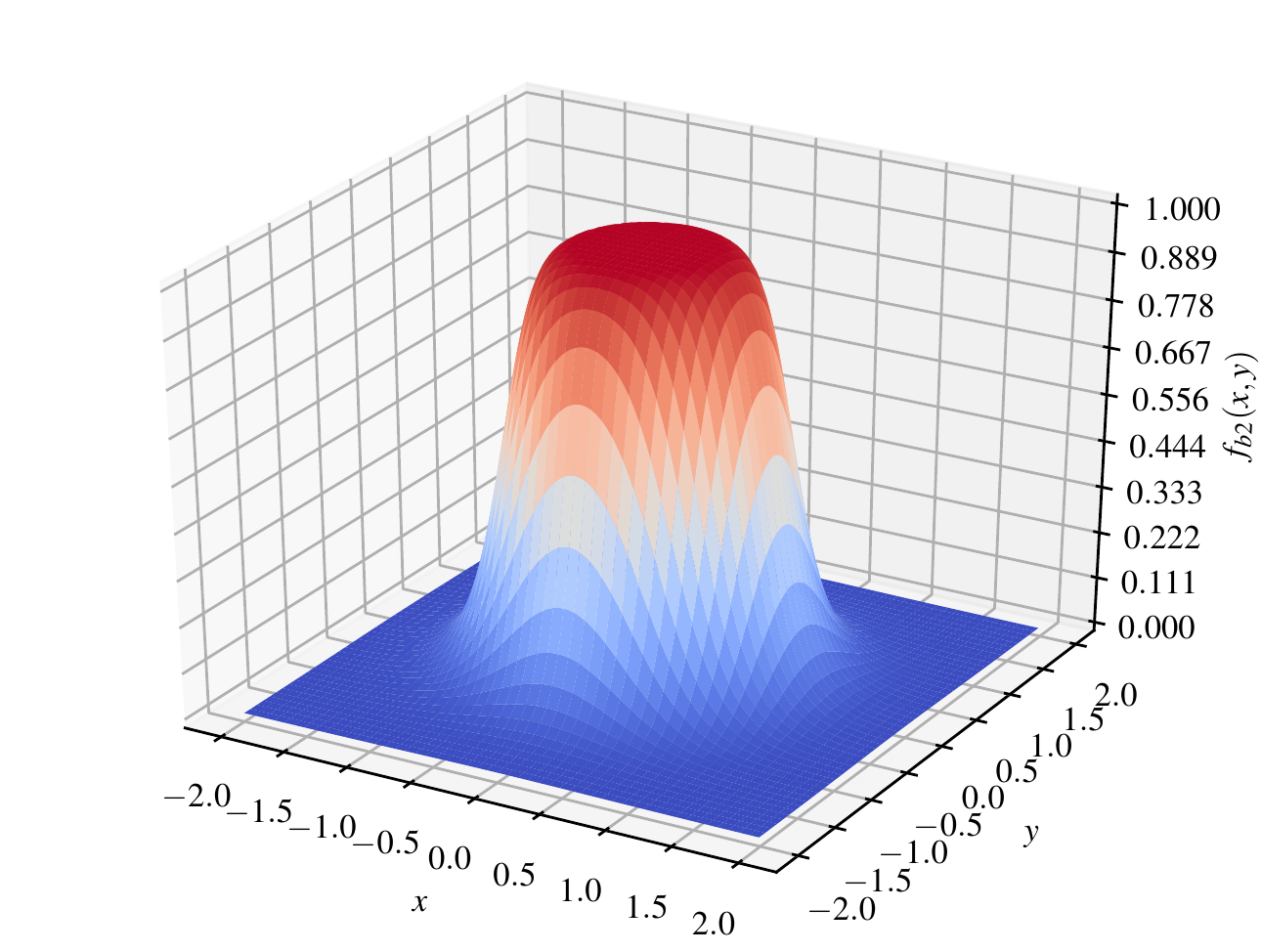}}
   \caption{
      (\textbf{a}) The Butterworth function~\cref{eq:butter} plotted for different values of $n_b$.
      The red vertical dashed line denotes the cut-off range parameter $c_b = 1$.
      The function satisfies $f_b(c_b) = 0.5$; therefore, the value of $c_b$ indicates an approximate sensor range.
      (\textbf{b}) The sensor range model is a function of two variables~\cref{eq:butter2}, plotted for $n_b = 8$ and $c_b=1$.}
   \label{fig:butterworth}
\end{figure}

The UAV sensor range, modeled as a radius around the vehicle where the reward is collected, is depicted in~\cref{fig:butterworth}.
The proposed sensor range model is based on the continuous differentiable Butterworth function~\cref{eq:butter} employed to serve as a reward collection indicator function in the reward model.

The degree $n_b$ of the Butterworth function influences the convergence rate in limits and function shape.
The value of $c_b$ specifies the effective sensor range as $f_b$ approximates square shape with increasing $n_b$.
It enables us to model the sensor coverage range as a function of the UAV's position~\cref{eq:butter2}.
\begin{align}
  f_b(x) &= \frac{1}{1 + (\frac{x}{c_b})^{n_b}}\label{eq:butter} \\ 
  f_{b2}(x,y) &= \frac{1}{1 + (\frac{\sqrt{x^2 + y^2}}{c_b})^{n_b}}
  \label{eq:butter2}
\end{align}

Due to the asymptotic convergence in $\lim\limits_{x\to\pm\infty} f_b(x) = 0$, the modeled reward collection is exhibited even for large distances $\gg c_b$; the model reward values are approximate.
It positively influences solution convergence of interior-point methods, as the value of $f_b$, which directly affects the reward gain, is always non-zero for all reward pairs of state-vehicle (up to the limits of the numerical precision).
On a UAV visit to the neighborhood of a rewarding target, the related reward state is approximately nullified due to $f_b(0) = 1$; the overall reward state error is assumed to converge to some finite value during the continual monitoring.

% - subsection ----------------------------------------------------------------
\subsection{Reward Model}

The dynamics of discrete reward state $r_i$, where $1 \leq i \leq n_p$, are modeled as a DAE~\cref{eq:dynamic_disc}, \reviewii{which is similar to the information accumulation function in~\cite{smith2012persistent}}.
\begin{align}\label{eq:dynamic_disc}
   r_{i,k+1} = & (r_i + T_s\Kgain)\nonumber\\
   &( 1 - \text{\text{max}}(1, \sum_{j=1}^{m} f_{b2}(p_{i,x,k} - q_{j,x,k}, p_{i,y,k} - q_{j,y,k}))
\end{align}
The reward variables' dynamics is autonomous if no vehicle is present proportional to $\Kgain$.
On the presence of a vehicle in the reward state neighborhood indicated by~\cref{eq:butter2}, the reward state is nullified.
We consider the sampling rate of the discrete dynamical system~$T_s$.
Limiting the maximal influence of the multiple vehicles on the reward states is necessary for the model, as the reward state values are assumed to be non-negative.

Due to the approximate sensor range radius $c_b$, the results of the proposed monitoring approach are processed by the evaluation model~\cref{eq:discrete_reward}.
Discrete vehicle positions $q_{j,x,k},p_{j,y,k}$ at sampled time-steps $k$ are used as the inputs of an indicator function to determine if a vehicle passed through the reward-collecting range during continual monitoring
\begin{align}
   &r_{i,k+1}^{\text{eval}} = 
   \begin{cases}
      0, & \text{if}\ d_{i,j} \leq n_{e},\ \forall 1 \leq j \leq m,\\
      r_{i,k}^{\text{eval}} + T_s\Kgain & \text{otherwise,}
   \end{cases}\label{eq:discrete_reward}\\
   & \text{where}\ d_{i,j} = \sqrt{(p_{i,x} - q_{j,x,k})^2 + (p_{i,y} - q_{j,y,k})^2}.\nonumber
\end{align}

% - subsection ----------------------------------------------------------------
\subsection{Vehicle Model}

The vehicles are modeled as the second-order DAE systems of $m$ multi-rotor UAV agents in~\cref{eq:multi_rotora,eq:multi_rotorb,eq:multi_rotorc,eq:multi_rotord} with the accelerations $a_{i,x}$, $a_{i,y}$ as the inputs.
The model is subject to motion constraints expressed in~\cref{eq:multi_rotor_cona,eq:multi_rotor_conb,eq:multi_rotor_conc}.

\vspace{-2em}
\begin{multicols}{2}\noindent
   \begin{subequations}\label{eq:multi_rotor}
      \begin{align}
	 &q_{i,x,k+1} = v_{i,x,k}T_s + a_{i,x,k}\frac{T_s^2}{2}\label{eq:multi_rotora}\\
	 &q_{i,y,k+1} = v_{i,y,k}T_s + a_{i,y,k}\frac{T_s^2}{2}\label{eq:multi_rotorb}\\
	 &v_{i,x,k+1} = a_{i,x,k}T_s\label{eq:multi_rotorc}\\
	 &v_{i,y,k+1} = a_{i,y,k}T_s\label{eq:multi_rotord}\\
	 &~~~\text{for } 1 \leq i \leq m.\nonumber
      \end{align}
   \end{subequations}
   \begin{subequations}\label{eq:multi_rotor_con}
      \begin{align}
	 & \sqrt{v_{i,x}^2 + v_{i,y}^2} \leq \Vmax\label{eq:multi_rotor_cona}\\
	 & -\Amax \leq a_{i,x} \leq \Amax\label{eq:multi_rotor_conb}\\
	 & -\Amax \leq a_{i,y} \leq \Amax\label{eq:multi_rotor_conc}\\
	 &~~~\text{for } 1 \leq i \leq m.\nonumber
      \end{align}
   \end{subequations}
\end{multicols}
\vspace{-1em}

Collision avoidance puts minimal mutual distance constraints~\cref{eq:min_dista,eq:min_distb} on the agents as:
\vspace{-0.5em}
\begin{subequations}\label{eq:min_dist}
   \begin{align}
      &d_{i,j} \geq \Dmin,\label{eq:min_dista} \\
      &d_{i,j} = \sqrt{(q_{i,x}-q_{j,x})^2 + (q_{i,y}-q_{j,y})^2},\label{eq:min_distb}\\
      &~~~~~~\text{for } 1 \leq i,j \leq m,\ i \neq j.\nonumber
   \end{align}
\end{subequations}

% - subsection ----------------------------------------------------------------
\subsection{\reviewi{Water Surface Model}}

\reviewi{The water surface area is modeled as a finite grid of target points.
The movement of dynamic debris on the surface is modeled as local periodic oscillations combined with a time-dependent drift, a simplified combination of the water currents' influence, sail effect, and Stokes drift.
Precise modeling of underlying causes might improve the movement prediction, but it is considered out of this letter's scope.
The model is expressed for each axis as simplified time-dependent function~\cref{eq:stokes} parameterized by the amplitude $A_{\text{p}}$, angular velocity $\omega_{\text{s}}$, and the drift velocity $v_{\text{d}}$.
\begin{equation}\label{eq:stokes}
  x_{\text{d}}(t) = t v_{\text{d}} + A_{\text{p}}\sin{(t \omega_{\text{p}})}
\end{equation}
The function parameters are assumed to be known but might vary during the monitoring for each target.
Then, parameter changes can be determined from the visual measurements during the adaptive monitoring.}

% - section -------------------------------------------------------------------
\section{Proposed MPC-based Solution for IMP-DR}\label{sec:mpc}

We propose to address the motivational problem of multi-vehicle dynamic water surface monitoring using the proposed IMP-DR formulation and planning solution method based on the MPC~\cref{eq:ocp}.
In the combined adaptive exploration and monitoring problem, we seek to maximize the information gained from the specified targets.
The information is quantified as a reward value associated with the targets.
As the reward values continually increase, the planning goal is to minimize the values present in the reward states of the modeled environment.

We propose employing an MPC-based control technique to solve the multi-vehicle planning problem, as it is suitable for addressing vehicle constraints and time-evolving reward dynamics.
The utilized Optimal Control Problem of the MPC on $N_s$ steps, where $N_s = \frac{T_h}{T_s}$, is stated as
\begin{align}\label{eq:ocp}
   \optmin_{x,u} = \sum_{k=0}^{N_{s}-1} & ( f_l(\matr{x}_k) + \Delta \matr{u}_k^T k_R \Delta \matr{u}_k ) + f_m(\matr{x}_{N_s}),\\
   \text{s.t.}\quad
   &\matr{x}_{lb} \leq \matr{x}_k \leq \matr{x}_{ub},\nonumber\\
   &\matr{u}_{lb} \leq \matr{u}_k \leq \matr{u}_{ub},\nonumber\\
   &\matr{z}_{lb} \leq \matr{z}_k \leq \matr{z}_{ub}.\nonumber
\end{align}
The state vector $\matr{x}$~\cref{eq:states} and input vector $\matr{u}$~\cref{eq:inputs} are:
\begin{align}
   \matr{x} = &
   \left[\begin{matrix} 
      q_{1,x},q_{1,y},q_{2,x},q_{2,y},\dots,q_{m,x},q_{m,y},
   \end{matrix}\right.\nonumber\\
   &\,\left.\begin{matrix}
      v_{1,x},v_{1,y},v_{2,x},v_{2,y},\dots,v_{m,x},v_{m,y},
   \end{matrix}\right.\nonumber\\
   &\,\left.\begin{matrix} 
      r_1,r_2\dots,r_{n_p}
   \end{matrix}\right]^T,\label{eq:states}\\
\matr{u} = &\begin{bmatrix} 
      a_{1,x},a_{1,y},a_{2,x},a_{2,y},\dots,a_{m,x},a_{m,y}
   \end{bmatrix}^T.\label{eq:inputs}
\end{align}
The functions $f_l(\matr{x}) = f_m(\matr{x}) = \sum_{k=1}^{n_p} r_k^2$ are the Lagrange and Meyer terms.
Input penalty $k_R = 10^{-3}$ is used to penalize the system input changes $\Delta \matr{u}_k = \matr{u}_{k+1} - \matr{u}_{k}$.
Including the penalty term prevents oscillatory behavior, and it was empirically observed to decrease solution times.
The constant vectors $\matr{x}_{ub}$, $\matr{x}_{lb}$, $\matr{u}_{ub}$, and $\matr{u}_{lb}$ enforce state and input constraints.
Algebraic vector $\matr{z}(\matr{x},\matr{u})$ and its bounding vectors $\matr{z}_{ub}$ and $\matr{z}_{lb}$ enforce minimum vehicle distance and velocity magnitude constraints.

% - section -------------------------------------------------------------------
\section{Results}\label{sec:results}

The proposed solution has been empirically examined in several scenarios to evaluate its properties.
In~\cref{sec:evaluation}, we present results on computational performance according to the problem size, number of vehicles, and planning horizon length.
Furthermore, the introduced IMP-DR generalizes both the Kinematic and Team OP. 
Therefore, the proposed approach is utilized to address these problems as well.
In particular, the proposed approach's performance is compared to the state-of-the-art Kinematic OP planner~\cite{Meyer2022} and Team OP meta-heuristic planner~\cite{pedersen2021grasp} with the results reported in \cref{sec:kinematicop} and \cref{sec:top}, respectively.
Finally, simulation results and reports on experimental field deployment on continual water surface exploration and monitoring tasks are presented in \cref{sec:water_surface_monitoring} and \cref{sec:experiment}, respectively.

% - subsection ----------------------------------------------------------------
\subsection{Computational Evaluation}\label{sec:evaluation}

For the computational evaluation, the static targets are sampled on a symmetric two-dimensional grid of the width $w_{\text{grid}}$.
The spacing of targets is $\SI{1}{\meter}$ on both axes if not otherwise specified.
The reward increase rate is set to $k_{gain} = \SI{1}{\per\second}$.
The initial reward state value is set to $r_{0} = 10$, as the system input change is penalized in the solver cost function.
The vehicle dynamics is limited by the maximum velocity vector magnitude $\Vmax = \SI{1}{\meter\per\second}$ and acceleration $\Amax = \SI{2}{\meter\per\second}$ in each axis.
The sensor model is set to $c_b = 0.25$ and $n_b = 8$.
The solutions were determined at each planning step with the period of $T_s=\SI{0.25}{\second}$.
The number of sampling steps in the planner is set to $N_s = 20$ with the resulting planning horizon of $\SI{5}{\second}$.
Monitoring was performed for $\SI{5}{\minute}$, translating to $S = \num{1200}$ planning steps.

The proposed approach is implemented using the \texttt{do-mpc} toolbox~\cite{dompc} with the \texttt{Ipopt}~\cite{ipopt} non-linear optimization framework.
The MA97 sparse linear system solver, part of the HSL collection~\cite{hsl_collection}, was utilized.
The evaluation was performed using \emph{Robot Operating System}~(ROS) control pipeline of the MRS~UAV~System~\cite{mrs_system} and the AMD Ryzen 4750U processor running at the base clock of \SI{1.7}{\giga\hertz}, accompanied with \SI{32}{\giga\byte} RAM.
The solution convergence tolerance of $10^{-8}$ was used as the stopping criterion during evaluation.

\begin{figure}[htbp]\centering
   \subfloat[]{\includegraphics[width=0.5\columnwidth]{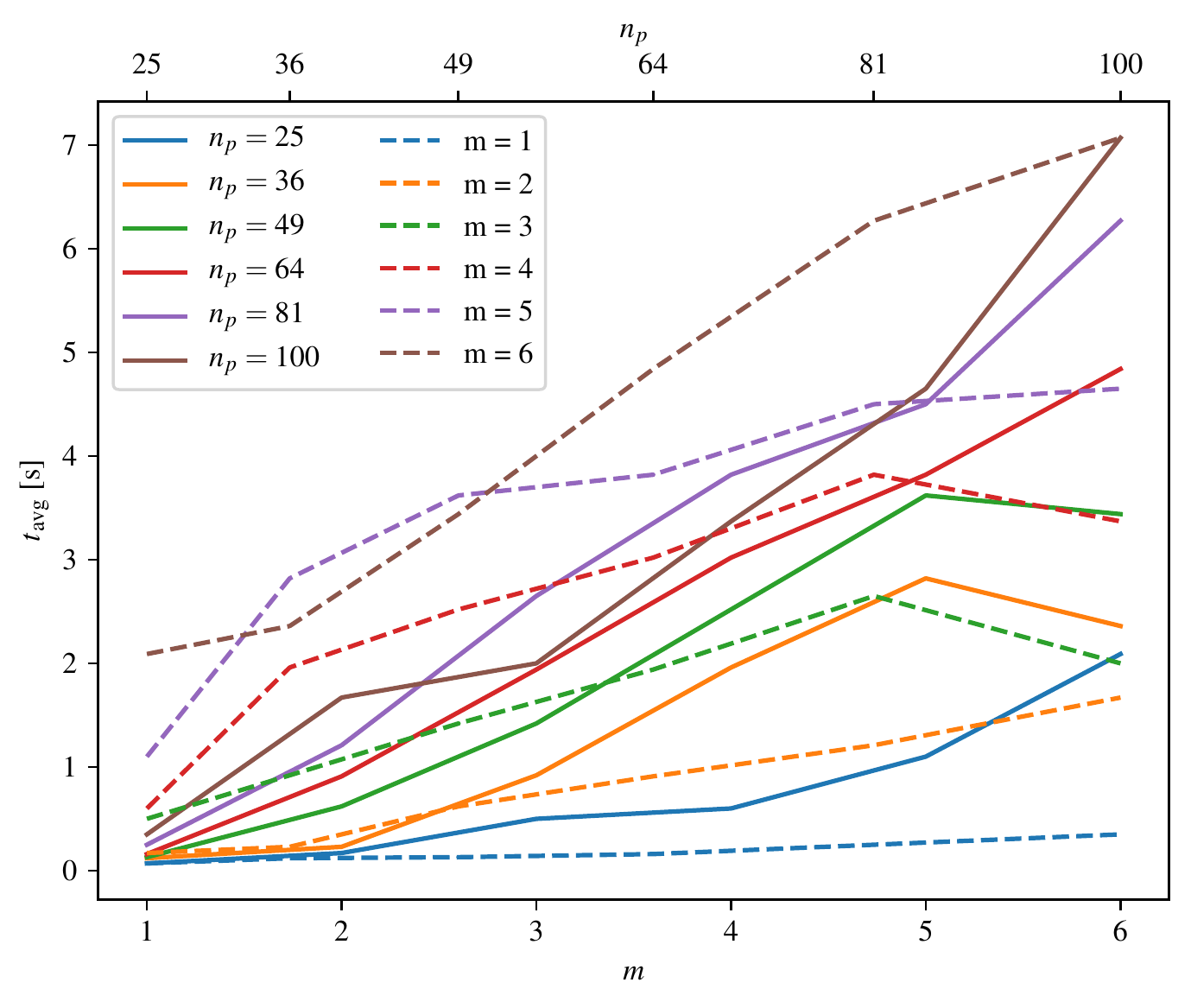}\label{fig:vehicles_width}}
   \hfill
   \subfloat[]{\includegraphics[width=0.5\columnwidth]{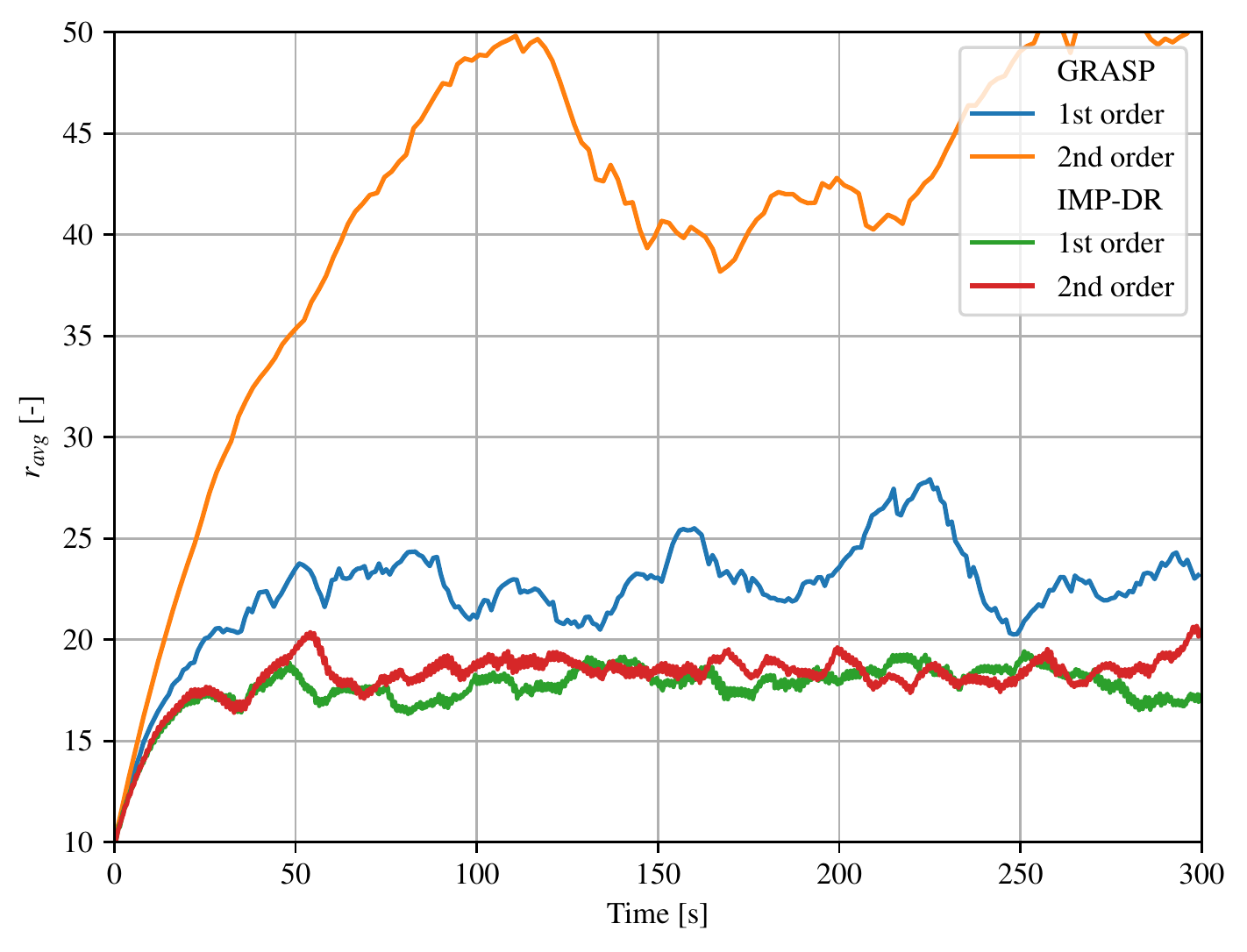}\label{fig:top_comparison}}
   \caption{\protect\subref{fig:vehicles_width} Average computational time $t_{\text{avg}}$ per planning step for the number of targets $n_p$ and $m$ vehicles.
   The time $t_{\text{avg}}$ does not necessarily increase with $n_p$ due to the utilization of multi-threading for larger problems by the MA97 solver; however, the overall tendency is linear with increasing problem size, as shown for $m=6$.
\protect\subref{fig:top_comparison} Average reward state values $r_{\text{avg}}$ in time of the GRASP-based TOP planner~\cite{pedersen2021grasp} and the proposed IMP-DR-based approach for the 1st order (velocity only) and the full 2nd order vehicle dynamics models.
      The monitoring problem is to minimize the system reward states by repeatedly visiting targets, nullifying the dynamic value.}
\end{figure}

\begin{table}[!htb]\centering
   \caption{Computational results for $n_p$ targets and $m$ vehicles}\label{tbl:vehicles_width}
   \vspace{-1em}
   %   \resizebox{\columnwidth}{!}{
\begin{tabular}{
      C
      C
      R
      R
      R
      R
      R
      R
   }
   \noalign{\hrule height 1.1pt}%\noalign{\smallskip}
   n_p & m & t_{\text{avg}}\text{ [\si{s}]} &t_{\text{max}} \text{ [\si{s}]}&r_{\text{eq}}\text{ [-]}& r_{\text{max}} \text{ [-]}\rule{0pt}{2ex}\\
   \midrule
   %\hline\noalign{\smallskip}
   25 & 1 & 0.05 & 1.5 & 32.7 & 109.8\\
25 & 2 & 0.12 & 2.0 & 27.6 & 139.2\\
25 & 3 & 0.55 & 5.8 & 16.7 & 97.7\\
25 & 4 & 0.77 &8.9 & \mathbf{9.9} & \mathbf{56.2}\\
25 & 5 & 1.06 & 8.8 & 13.7 & 94.6\\
25 & 6 & 1.22 & 12.1 & 16.4 & 86.2\\
\hline\noalign{\smallskip}
49 & 1 & 0.18 & 5.8 & 31.3 & 125.0\\
49 & 2 & 0.49 & 4.3 & 32.3 & 143.5\\
49 & 3 & 1.42 & 19.9 & 18.3 & 80.6\\
49 & 4 & 1.63 & 21.1 & 25.5 & 98.5\\
49 & 5 & 3.30 & 18.3 & 14.4 & 94.6\\
49 & 6 & 3.53 & 33.0 & \mathbf{13.3} & \mathbf{70.0}\\
\hline\noalign{\smallskip}
100 & 1 & 0.43 & 20.9 & 55.2 & 279.9\\
100 & 2 & 1.50 & 8.9 & 33.4 & 149.0\\
100 & 3 & 1.85 & 26.9 & 28.1 & 164.1\\
100 & 4 & 2.67 & 28.5 & 28.2 & 118.9\\
100 & 5 & 4.16 & 31.8 & 24.5 & 136.5\\
100 & 6 & 7.07 & 77.3 & \mathbf{22.7} & \mathbf{118.4}

   \\\noalign{\hrule height 1.1pt}
\end{tabular}
%}

\end{table}

The influence of the number of vehicles $m$ and targets $n_p$ on the computational performance is depicted in \cref{fig:vehicles_width} and listed in~\cref{tbl:vehicles_width}.
Since the computational time differs at each planning step, the average $t_{\text{wall}}$ planning time is reported as $t_{\text{avg}}$ and the maximal planning time as $t_{\text{max}}$.
The average reward state value $r_{\text{avg}} = \frac{1}{n_p}\sum_{i=1}^{n_p}r_i$ is continually stable around the equilibrium value $r_{\text{eq}}$, the maximum system state reward is reported in the column $r_{\text{max}}$ in \cref{tbl:vehicles_width}.
From the observations of $r_{\text{avg}}$, its value exhibits Lyapunov stability as it does not diverge from a neighborhood of $r_{\text{eq}}$ for $t\to\infty$ that has been pragmatically examined for $\SI{300}{\second}$ as an average of $r_{\text{avg}}$.
It demonstrates the numerical stability of the proposed continual monitoring approach in indefinite operation.
The average step computational time $t_{\text{avg}}$ increases linearly with the number of targets $n_p$ and polynomially with the number of vehicles~$m$.

\begin{table}[htbp]\centering
   \caption{Computational results for $m$ vehicles and horizon~$N_s$}\label{tbl:horizon}
   \vspace{-1em}
   %\resizebox{\columnwidth}{!}{
\begin{tabular}{
      C
      C
      R
      R
      R
      R
      R
      R
   }
   \noalign{\hrule height 1.1pt}%\noalign{\smallskip}
   m & N_s & t_{\text{avg}}\text{ [\si{s}]} &t_{\text{max}} \text{ [\si{s}]}&r_{\text{eq}}\text{ [-]}& r_{\text{max}} \text{ [-]}\rule{0pt}{2ex}\\
   \hline\noalign{\smallskip}
   1 & 10 & 0.05 & 0.4 & 78.4 & 309.7\\
1 & 15 & 0.16 & 11.9 & 64.9 & 268.2\\
1 & 20 & 0.34 & 2.8 & 61.0 & 237.5\\
1 & 25 & 0.61 & 6.0 & 59.3 & 299.2\\
1 & 30 & 1.12 & 9.9 & \mathbf{54.7} & \mathbf{255.0}\\
\hline
3 & 10 & 0.23 & 2.3 & 71.0 & 294.9\\
3 & 15 & 1.32 & 19.3 & 28.6 & 191.8\\
3 & 20 & 1.63 & 10.4 & 28.3 & 155.5\\
3 & 25 & 3.17 & 17.6 & 26.6 & 138.7\\
3 & 30 & 5.39 & 33.2 & \mathbf{26.2} & \mathbf{135.0}\\
\hline
5 & 10 & 0.5 & 11.1 & 52.5 & 279.2\\
5 & 15 & 1.61 & 9.4 & 23.4 & 153.4\\
5 & 20 & 4.29 & 26.5 & 20.4 & 133.3\\
5 & 25 & 11.02 & 84.5 & \mathbf{17.1} & \mathbf{100.9}\\
5 & 30 & 21.68 & 224.7 & 21.2 & 137.7

   \\\noalign{\hrule height 1.1pt}
\end{tabular}
%}

\end{table}
The influence of the planning horizon $N_s$ on the computational performance with $w_{\text{grid}}=10$ is depicted in \cref{tbl:horizon}.
While increasing $N_s$ yields an exponential increase in $t_{\text{avg}}$, it does not necessarily lead to improved average reward equilibrium $r_{\text{eq}}$ and the maximal reward $r_{\text{max}}$.
Hence, lower values of $N_s$ can be preferred.

Here, it is worth noting that the used \texttt{Ipopt} solver relies on the third-party code to solve sparse symmetric indefinite linear systems repeatedly, and the choice of the third-party solver influences the required computational time and the quality of the solution.
A solution is needed in less than the sampling rate $T_s$ for the ideal operation of the proposed monitoring approach.
Although planning at each sampling step provides the best performance, it can be performed at a reduced rate to allow for a longer computational time.
Besides, the solver can be terminated prematurely before its convergence to obtain an intermediate solution suitable for field deployment, which is used for the results presented in \cref{sec:experiment}.

% - subsection ----------------------------------------------------------------
\subsection{Kinematic Orienteering Problem (Kinematic OP)}\label{sec:kinematicop}

The IMP-DR problem can be formulated as the Kinematic OP if the reward dynamics is non-existent and end-point constraints are introduced.
The proposed MPC-based approach can provide an approximate solution to the single-vehicle Kinematic OP formulated in~\cite{Meyer2022}.
Therefore, the proposed IMP-DR is compared with the state-of-the-art solutions to the Kinematic OP denoted KOP-1 and KOP-6\textsuperscript{lns}~\cite{Meyer2022}.

For the comparison, the OP is addressed by modeling the neighborhood size negligible to the overall target distances in the problem instance.
The sampling of $T_s = \SI{0.1}{s}$ was used with the input penalties of $k_Q = 10^{-3}$.
The cost function terms were modified as: $f_l(\matr{x}) = \sum_{k=1}^{n_p} r_k$, $f_m(\matr{x}) = 10^{3} ((q_x - x_{\text{final}})^2 + (q_y - y_{\text{final}})^2)$ to apply a quadratic soft-constraint on the final vehicle position $q_{N_s,x}, q_{N_s,y}$.
The sensor parameters were set to $c_b = 0.05$ and $n_b = 2$, approximating a spiking function at each target position.
Dynamical constraints were $\Vmax=\SI{3}{\meter\per\second}$ and $\Amax=\SI{1.5}{\meter\per\square\second}$.

\begin{table}[htb]\centering
   \caption{Results for the Kinematic Orienteering Problem}\label{tbl:kop}
   \vspace{-1em}
   \begin{tabular}{
      R
      R
      R
      R
      R
      R
   }
   \noalign{\hrule height 1.1pt}\noalign{\smallskip}
   %C_{\text{max}} [\text{s}]&\text{KOP-1}\cite{Meyer2022} & \text{KOP-6}^{\text{lns}}~\cite{Meyer2022} & \text{MPC} & t_{\text{avg}} [\text{s}]\\
   C_{\text{max}} \text{ [s]}&\text{KOP-1} & \text{KOP-6}^{\text{lns}} & \text{MPC}_{\text{best}} & t_{\text{avg}} \text{ [s]}\\
   \noalign{\smallskip}\hline\noalign{\smallskip}
   10 & 95 & 80 & \mathbf{115} &1.7 \\
   15 & 180 & 165 & \mathbf{200}&2.8 \\
   20 & 250 & 250 & \mathbf{260}&8.7 \\
   25 & 325 & 330 & \mathbf{370}&11.4 \\
   30 & 390 & 390& \mathbf{450}&15.9 \\
   35 & 430 & 435 & \mathbf{450}&22.8 \\
   40 & \mathbf{450} & \mathbf{450} & \mathbf{450}&45.7\\
   \noalign{\hrule height 1.1pt}
\end{tabular}
%}

\end{table}

\begin{figure}[htbp]\centering
   \subfloat[$C_{\text{\text{max}}}=\SI{10}{\second}$\label{fig:kop1}]{\includegraphics[width=0.5\columnwidth]{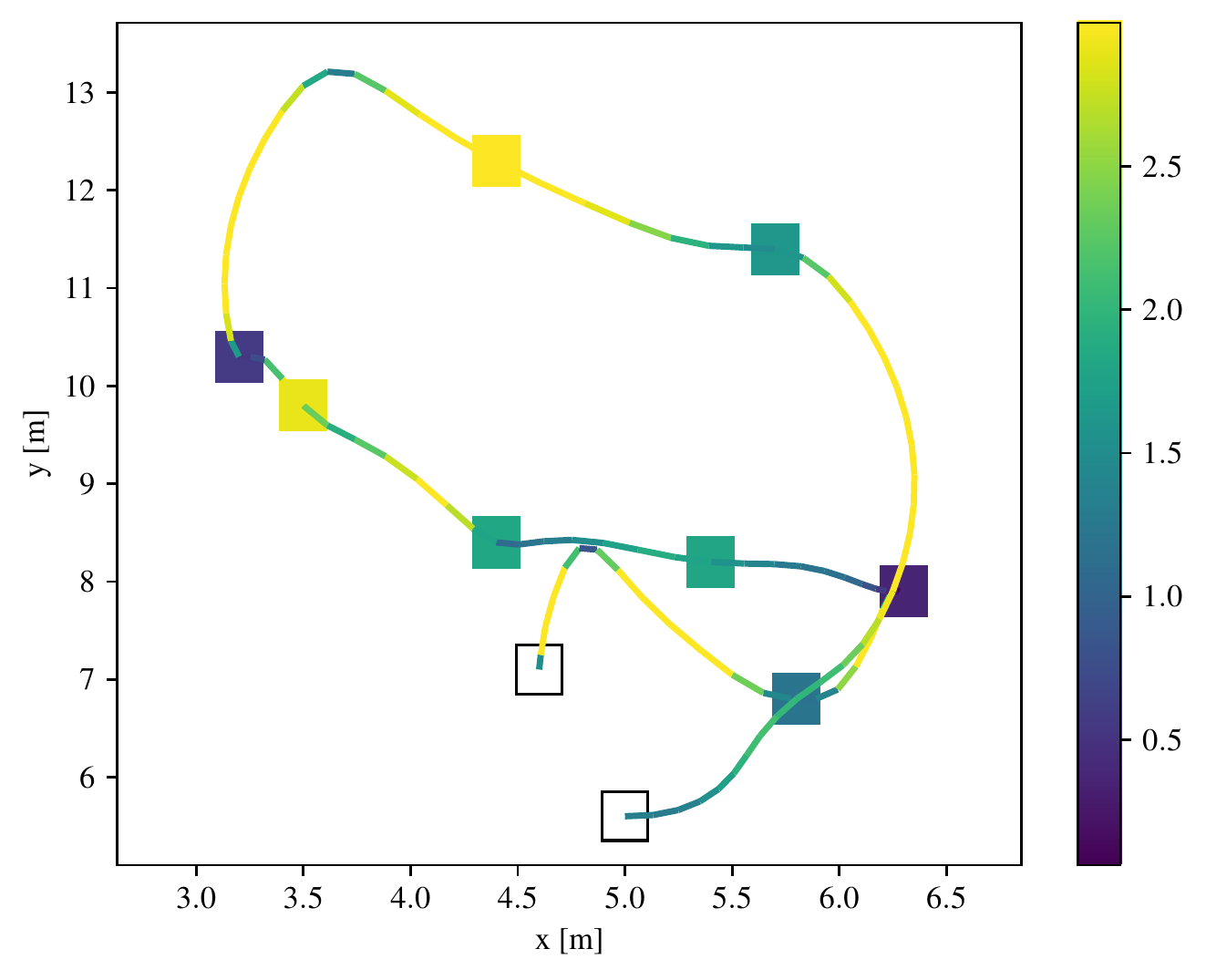}}
   \hfill
   \subfloat[$C_{\text{\text{max}}}=\SI{40}{\second}$\label{fig:kop2}]{\includegraphics[width=0.5\columnwidth]{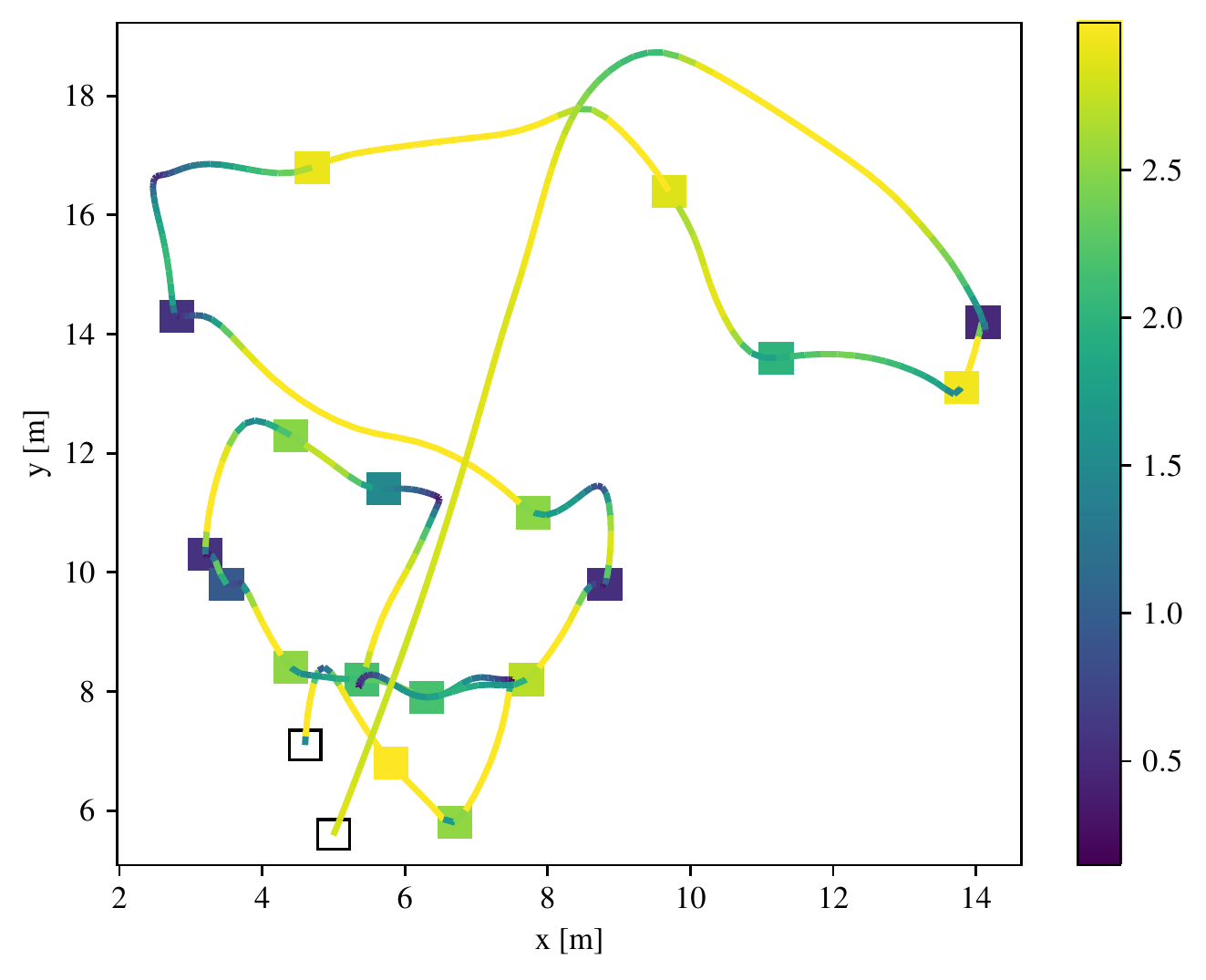}}
   \caption{Velocity profiles of two solutions to the Kinematic OP on the Tsigilirides-2 problem instances with the travel budget $C_{\text{\text{max}}}$ along the traveled path.
   The path color indicates the velocity for the scale depicted at the right of the plot in \si{\meter\per\second}.
   The solutions were obtained using the proposed MPC-based reward collection approach adjusted for solving the Kinematic OP.}
   \label{fig:kop}
\end{figure}

Furthermore, since the \texttt{Ipopt} solver provides a locally optimal solution, a potentially better solution can be obtained by perturbing the initial reward values in the dynamic model.
Thus, a noise with $\mathcal{N}(0,\,0.1)$ distribution was added to the initial reward values on each solution iteration, and \num{10} iterations were performed for each benchmark instance.
The best-found results are presented in~\cref{tbl:kop} under the column MPC\textsubscript{best} and shown in~\cref{fig:kop}.
The average computational time of the proposed MPC-based solver is denoted $t_{\text{avg}}$.

The proposed solution method provided improved reward gain in all but one testing instance, where the travel budget $C_{\text{\text{max}}}$ $[\si{s}]$ is sufficiently large for the compared methods to catch up with the proposed approach performance.
The trajectory deviation from the visited targets and final position does not exceed $\SI{0.02}{m}$ in all benchmark instances, which we consider negligible with regard to the problem scale.
As the number of planning steps is fixed, the provided solutions are not time-optimal.
They, however, satisfy the travel budget constraint.
In addition to better performance than the existing solvers in the collected reward, the proposed MPC-based approach can further address the reward dynamics, neighborhood, and multiple vehicles.

% - subsection ----------------------------------------------------------------
\subsection{Team Orienteering Problem (Team OP)}\label{sec:top}

A variant of the multi-vehicle monitoring scenario can be formulated as the Team OP, for which the proposed solver can also be utilized.
Therefore, we compared its performance with the grid-based Team OP planner proposed in~\cite{pedersen2021grasp}.
The \emph{Greedy Random Adaptive Search Procedure} (GRASP) meta-heuristic planner~\cite{pedersen2021grasp} determines paths to maximize the collected reward gain from the sampled grid locations using multiple vehicles with the given travel budget.
Hence, it was generalized for use in the receding horizon planning scenario by removing the end-point constraint and using individual vehicle starting locations. 
Besides, the GRASP cost function was modified to $f_l(\matr{x}) = \sum_{i=1}^{n_p}r_i^2$.
It provides a trade-off between the computational requirements and solution quality compared to the exact \emph{Mixed Integer Linear Programming} solution.
The computational effectiveness is essential for the real-time operation of the proposed monitoring approach, where the Team OP is solved repeatedly on each sampling step.
Therefore, we consider the heuristic approach in the comparison.

Grid width $w_{\text{grid}}=10$ with $n_p = 100$ and $m=3$ vehicles were used.
The GRASP planner was limited to movement in the 4-neighborhood of a two-dimensional grid at $\SI{1}{\meter\per\second}$.
Neglecting the second-order dynamics and utilizing maximal vehicle velocity provides a \emph{Lower Bound} (LB) on the GRASP-based monitoring performance. 
Assuming the vehicles have to accelerate and decelerate between the targets at $\SI{2}{\meter\per\second\squared}$ provides an \emph{Upper Bound} (UB) solution cost.
As the second-order dynamics become negligible (due to large target distances), the UB converges to the LB.

Five GRASP planning iterations were performed on each planning step to obtain a quality solution.
The planning was performed on the horizon of the 20 steps, i.e., \SI{20}{s} for the LB and $\SI{40}{s}$ for the UB.
The tuning of the proposed IMP-DR solver was the same as for the Kinematic OP reported in \cref{sec:kinematicop} with the sampling rate $T_s = \SI{0.1}{s}$, input penalties $k_Q = 10^{-3}$, neighborhood $c_b = 0.05$, $n_b = 2$, and $N_s = 20$ planning steps.
Solutions were obtained addressing the first-order velocity dynamics and the second-order acceleration dynamics vehicle models.
The benchmark was run for $\SI{300}{s}$, and the average system state reward values $r_{\text{avg}}$ provided by the evaluation model are plotted in~\cref{fig:top_comparison}.
On the target visit, vehicles did not exceed the neighborhood of $\SI{0.02}{m}$ as in the KOP benchmark.

The proposed IMP-DR planning approach demonstrates a~lower average reward $r_{\text{avg}}$ in continual monitoring over the GRASP-based TOP planner.
The IMP-DR planning performance is similar for both the first- and acceleration-constrained second-order models.
The results demonstrate the importance of the dynamic reward model in continual monitoring as it leads to lower $r_{\text{eq}}$.
However, IMP-DR is impractical for planning multi-vehicle information-gathering tasks with large travel budgets, as the $t_{\text{avg}}$ increases exponentially with planning horizon length. 

% - subsection ----------------------------------------------------------------
\subsection{Water Surface Monitoring and Flotsam Monitoring}\label{sec:water_surface_monitoring}

The addressed motivational scenarios of repeated water surface exploration to detect objects of interest and dynamic object monitoring are presented.
In the formulated exploration task, a symmetric target grid of $w_{\text{grid}}=20$ with the target spacing of $\SI{0.5}{m}$ was used.
The sensor function was set to $c_b = 0.5$ and $n_b = 4$, modeling the reward collection dependent on the distance.
The MPC parameters were set to $T_s=\SI{0.25}{s}$ and $N_s=20$.
\begin{figure}[!htbp]\centering
   \subfloat[$t=\SI{10}{s}$]{\includegraphics[width=0.5\columnwidth]{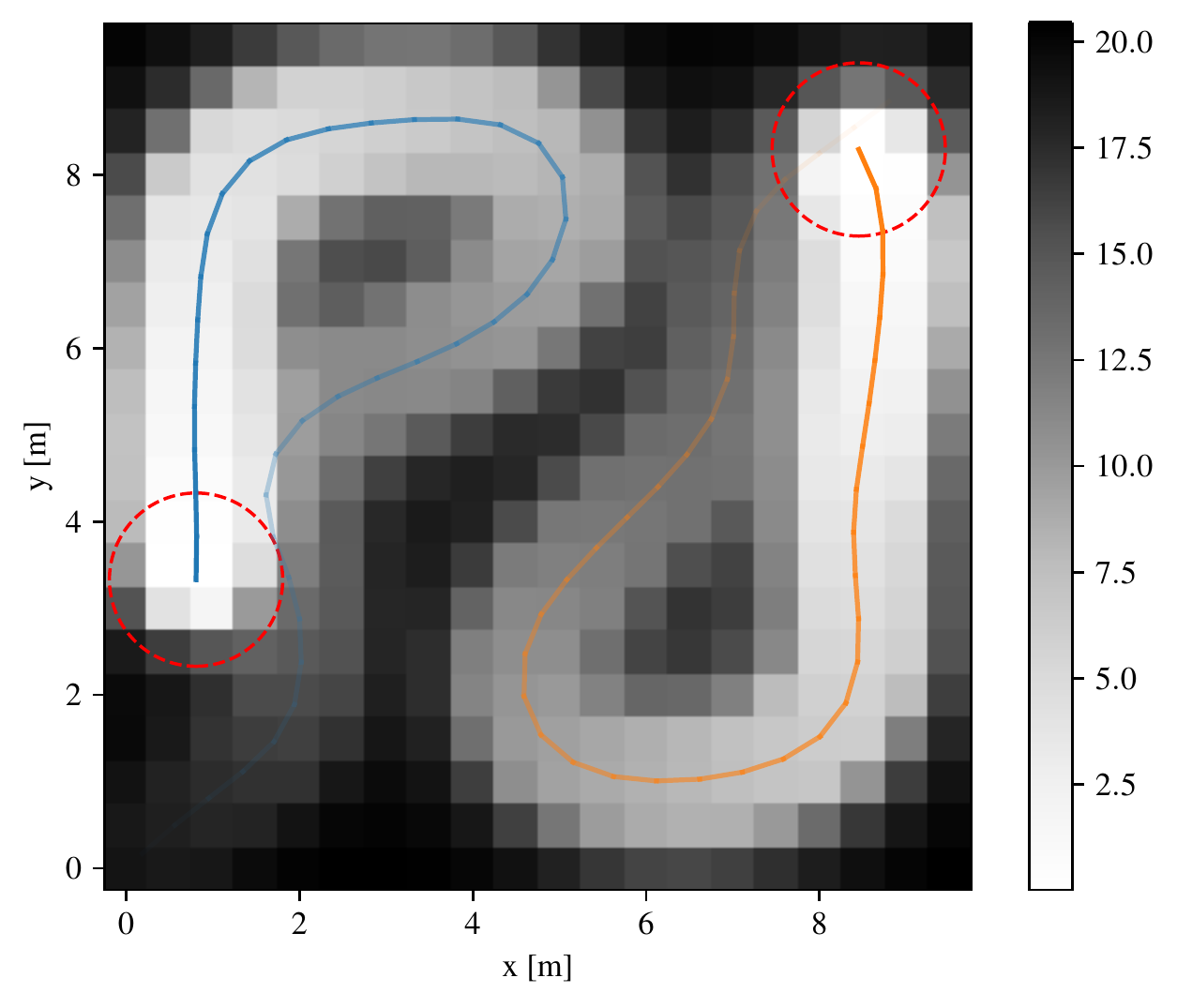}}
   \hfill
   \subfloat[$t=\SI{15}{s}$]{\includegraphics[width=0.5\columnwidth]{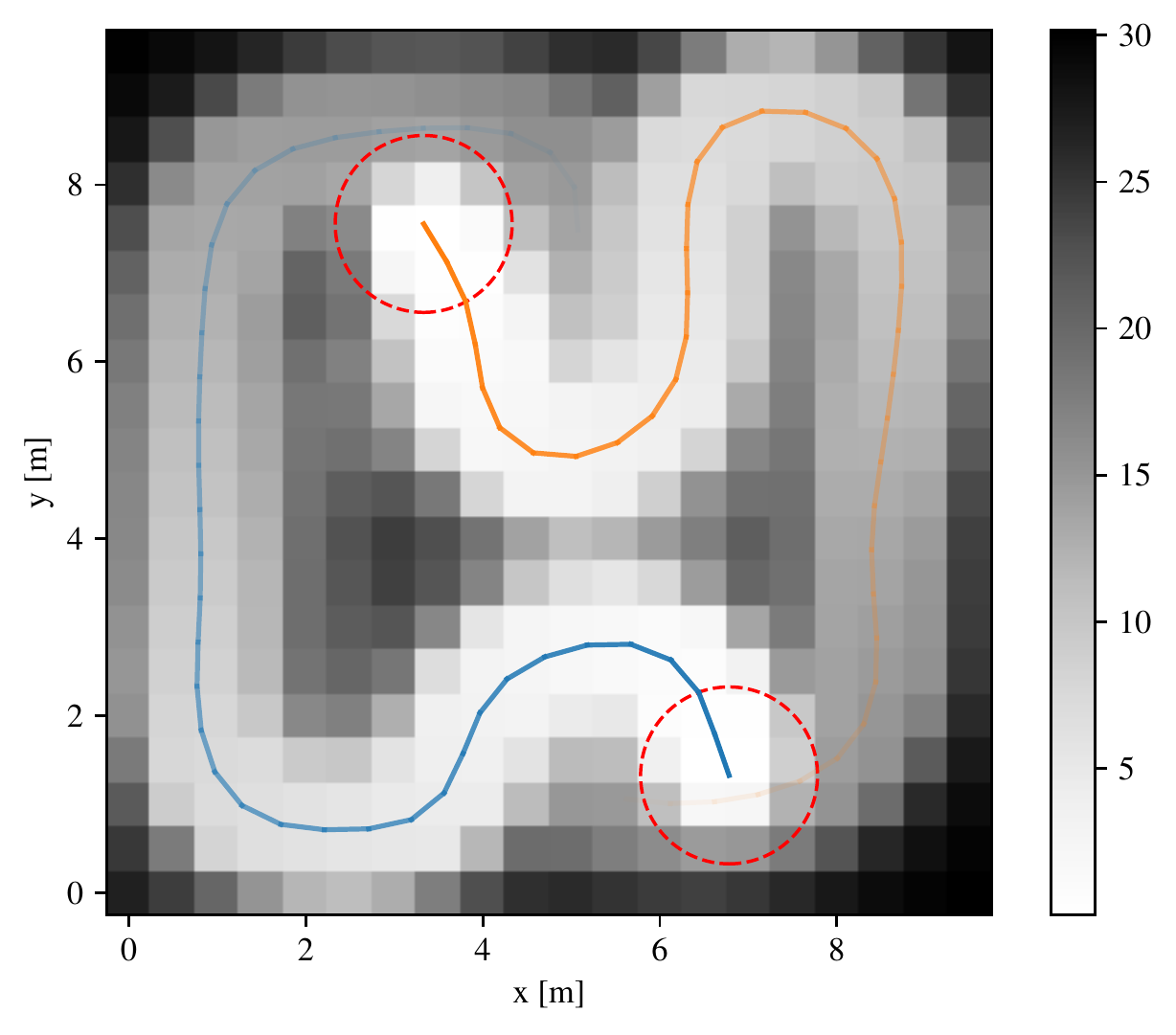}}

  \subfloat[$t=\SI{20}{s}$]{\includegraphics[width=0.5\columnwidth]{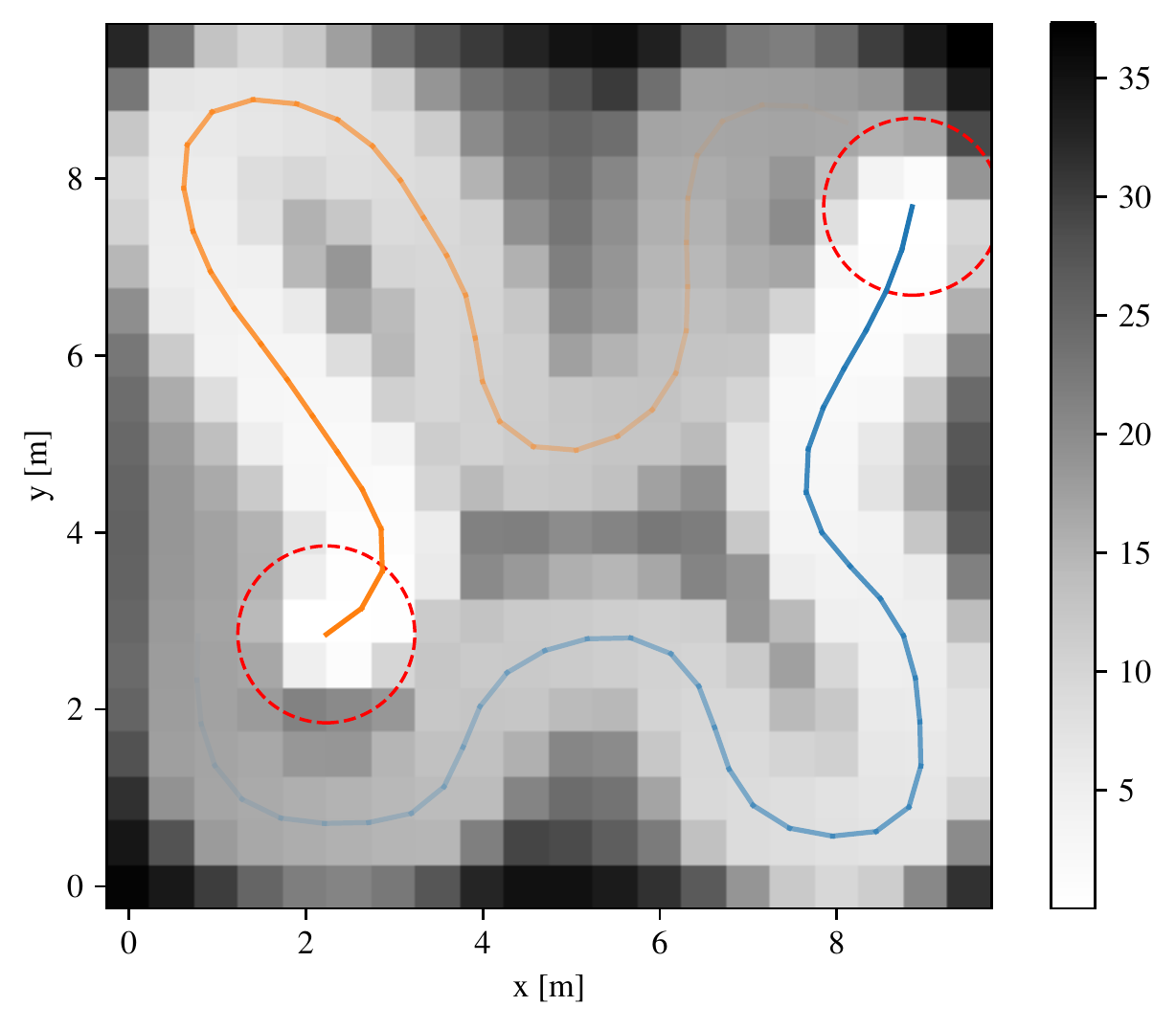}}
   \hfill
  \subfloat[$t=\SI{25}{s}$]{\includegraphics[width=0.5\columnwidth]{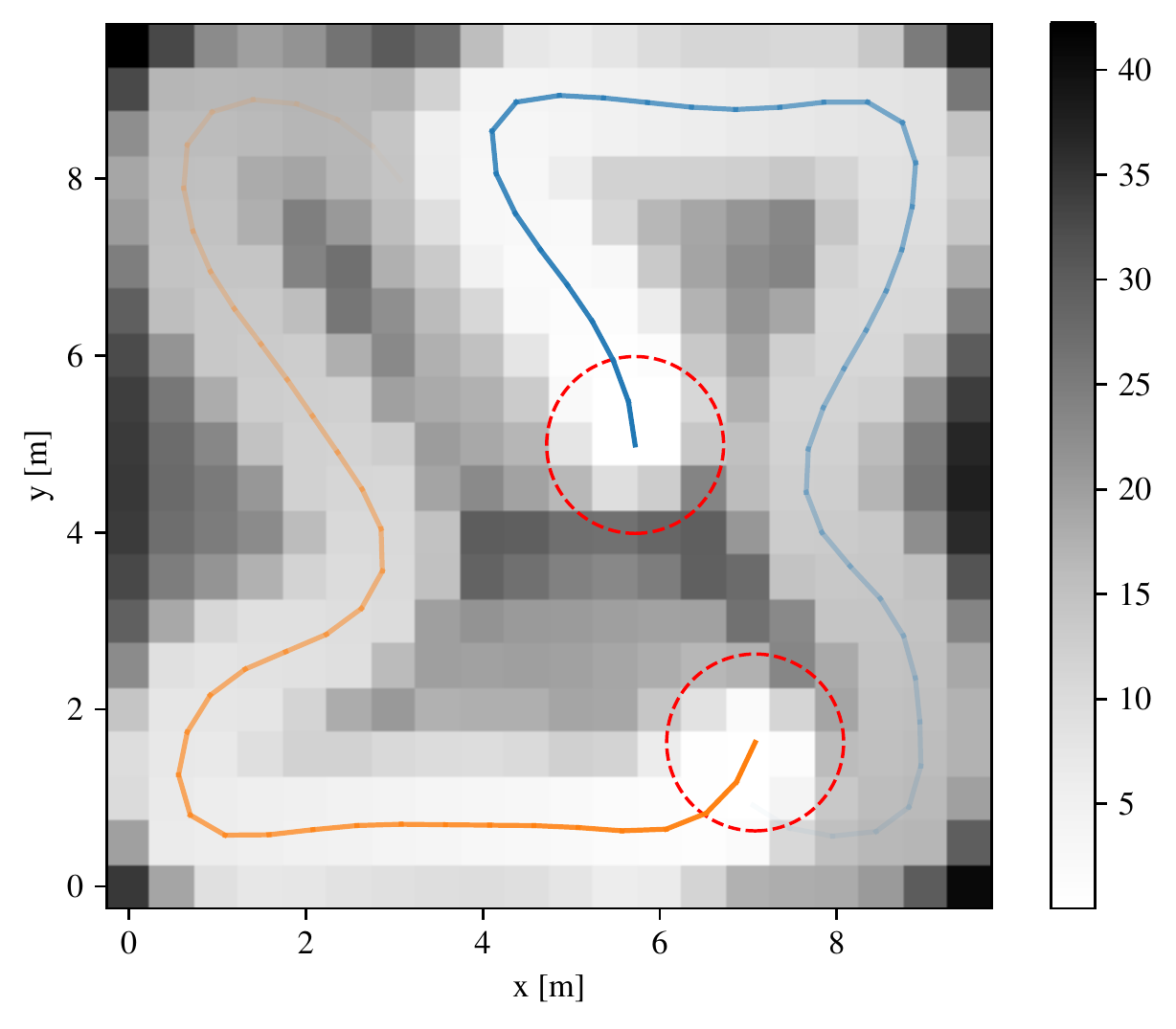}}
  \caption{Water surface monitoring scenario on $20\times20$ large grid utilizing two vehicles.
   The color of the cells indicates the reward state value according to the scale depicted at the right part of the plot.
   A circle with the radius $\SI{1}{\meter}$ is shown in the red around each vehicle at the time instant $t$, approximating the maximal reward collection range.
   The paths traveled by vehicles are in blue and orange.
  \label{fig:exploration}
  }
\end{figure}
The evolution of the system state rewards at a particular time instant $t$ is depicted in \cref{fig:exploration} together with the visualization of the vehicles' trajectories.
The task was experimentally evaluated, and results are reported in~\cref{sec:experiment}.

\begin{figure}[htbp]\centering
   \subfloat[$t =\SI{10}{s}$]{\includegraphics[width=0.5\columnwidth]{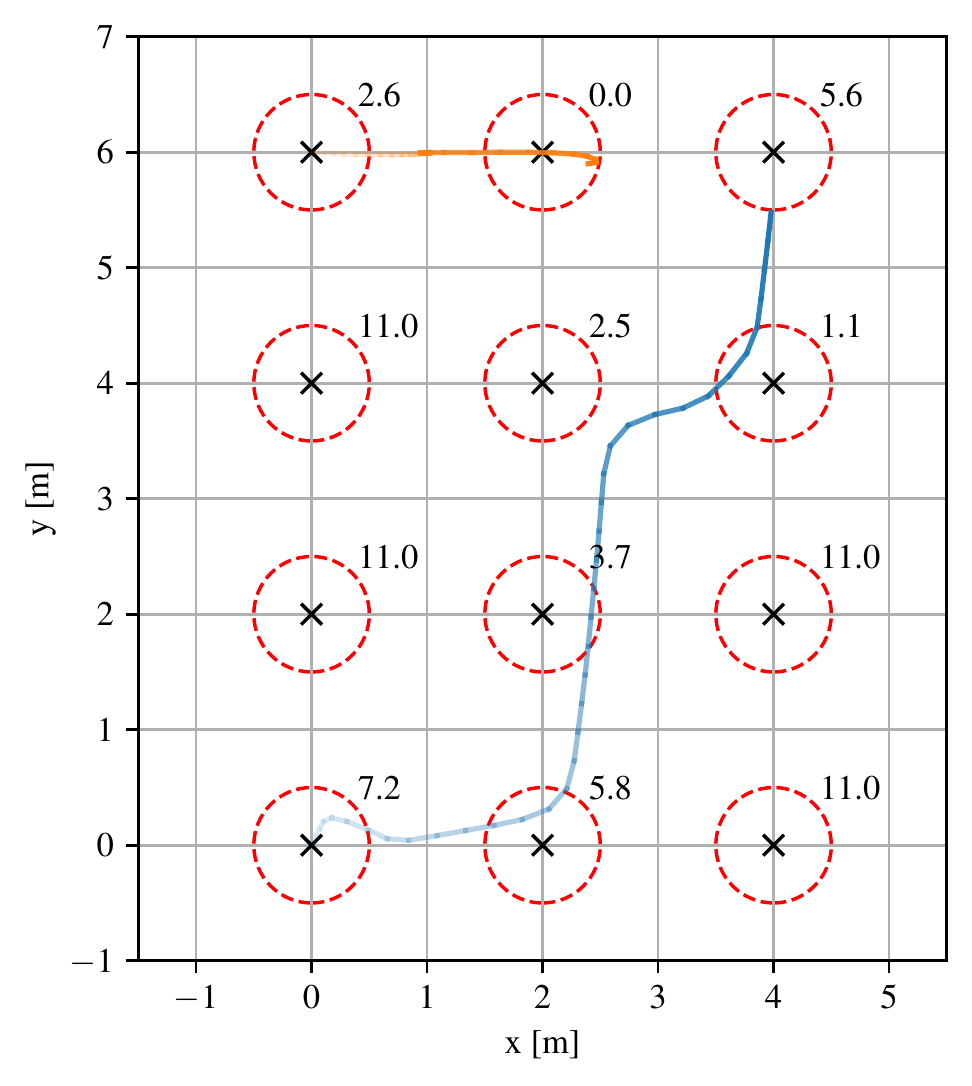}\label{fig:flotsam_a}}
   \hfill
   \subfloat[$t =\SI{15}{s}$]{\includegraphics[width=0.5\columnwidth]{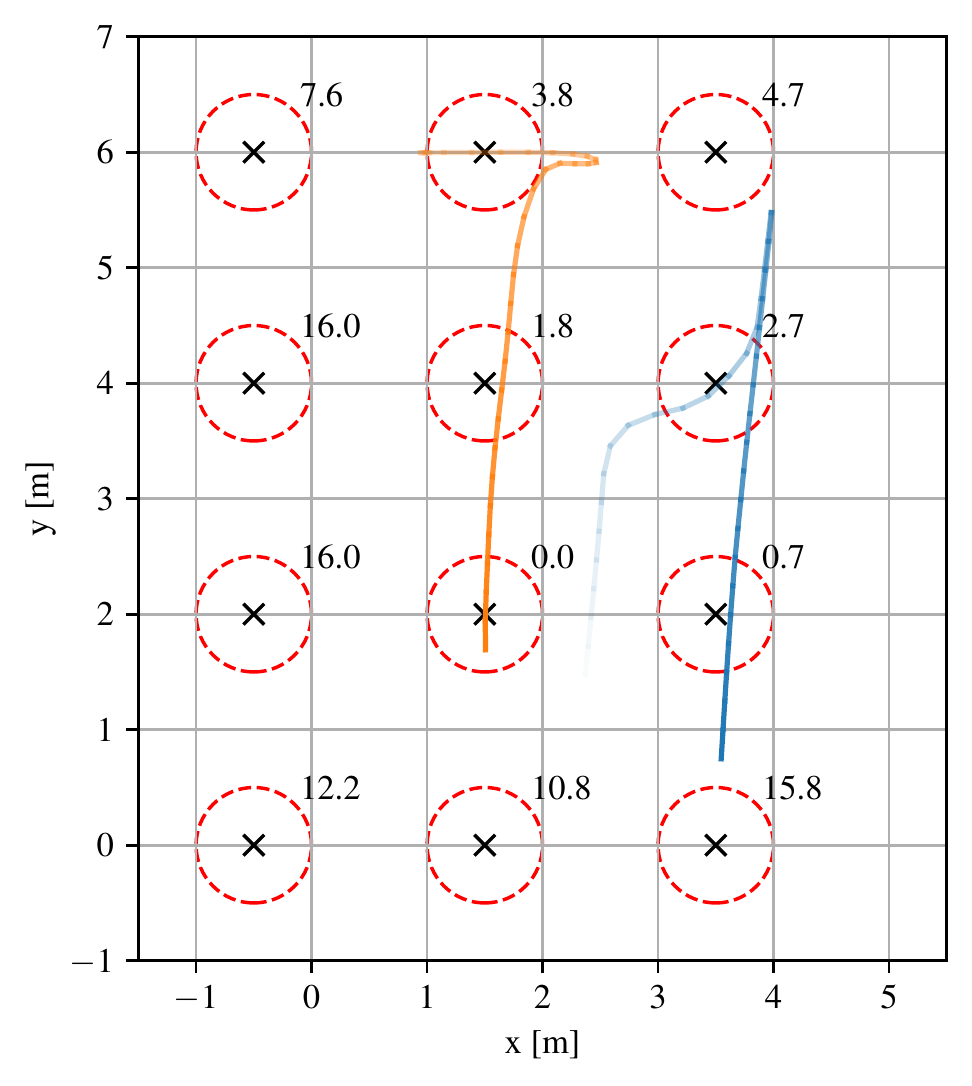}}
   \vspace{-0.5em}

   \subfloat[$t =\SI{20}{s}$]{\includegraphics[width=0.5\columnwidth]{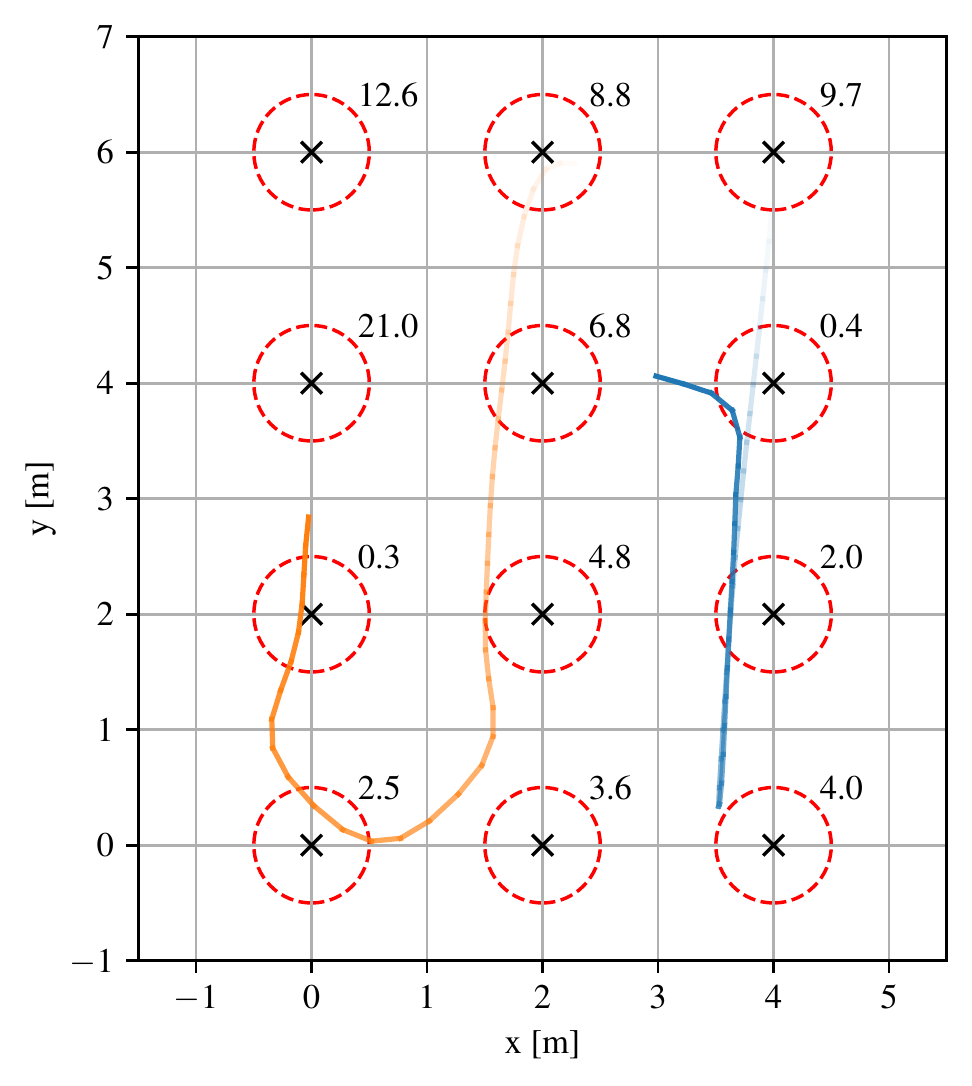}}
   \hfill
   \subfloat[$t = \SI{25}{s}$]{\includegraphics[width=0.5\columnwidth]{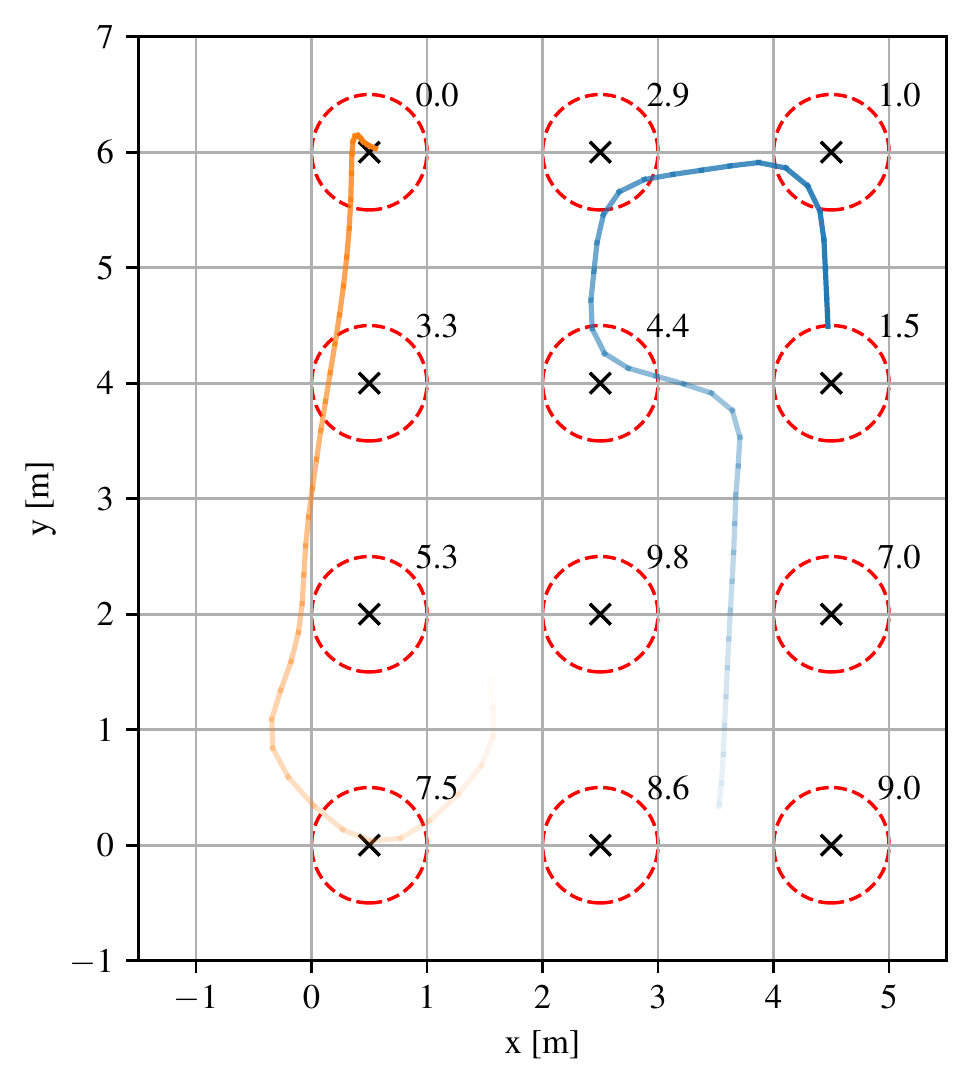}}
   \caption{Flotsam monitoring scenario with dynamic target positions and two vehicles. 
   The red (dashed) circles denote the reward collection radius of \SI{0.5}{m}.
   Paths traveled by vehicles are shown in blue and orange.
   }
   \label{fig:flotsam}
\end{figure}

Regarding the motivational scenario of monitoring debris or flotsam on a water surface with predictable movement dynamics, we consider the following problem setup.
The targets are within a $3\times4$ large grid with the spacing of $\SI{2}{m}$ and the exact vehicle dynamic constraints as in the previous water surface monitoring scenario.
\reviewi{The target dynamics in the x-axis are modeled as~\cref{eq:stokes} with $A_{\text{p}} = 0.5$, $\omega_{\text{p}}=\frac{\pi}{10}$, and $v_{\text{d}}=0$.}
Modeling target movement is possible by introducing time-varying parameters in the MPC planner.
The parameters for the scenario instance are set to $T_s=\SI{0.25}{s}$ and $N_s=40$, and the sensor function to $c_b = 0.5$ and $n_b = 8$.

The monitoring for the time horizon of $\SI{25}{s}$ is shown in~\cref{fig:flotsam}.
\reviewi{The vehicle positions are initiated from the opposite left-most targets and proceed to monitor moving targets continuously.
Note that the locally optimal solutions provided by the utilized solver tend to degrade if vehicles are in close proximity, as seen in~\cref{fig:flotsam_a}, where one of the vehicles is stuck in a feasible solution space.}

% - subsection ----------------------------------------------------------------
\subsection{Experimental Field Deployment}\label{sec:experiment}

\begin{figure}[htbp]\centering
   \subfloat[F450 UAV platforms\label{fig:f450}]{\includegraphics[width=0.45\columnwidth]{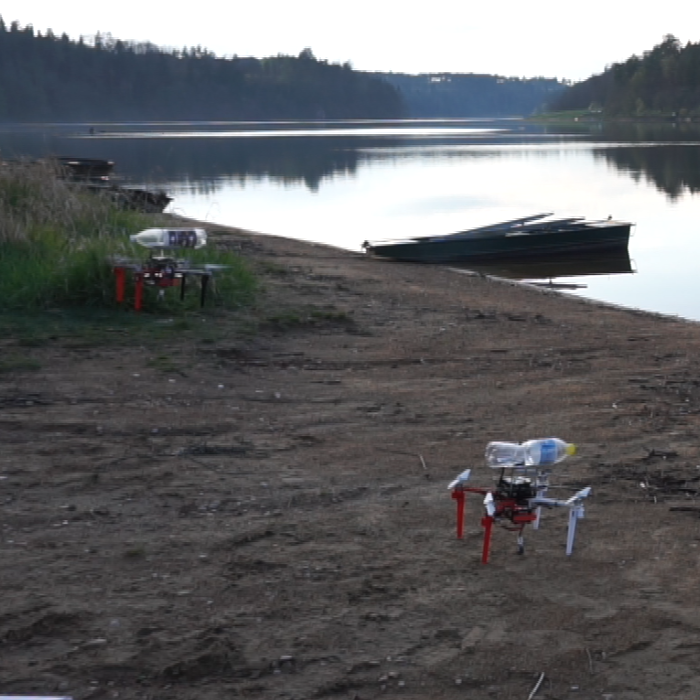}}
   \hfill
   \subfloat[Top-down view of the UAVs\label{fig:topdown}]{\includegraphics[width=0.45\columnwidth]{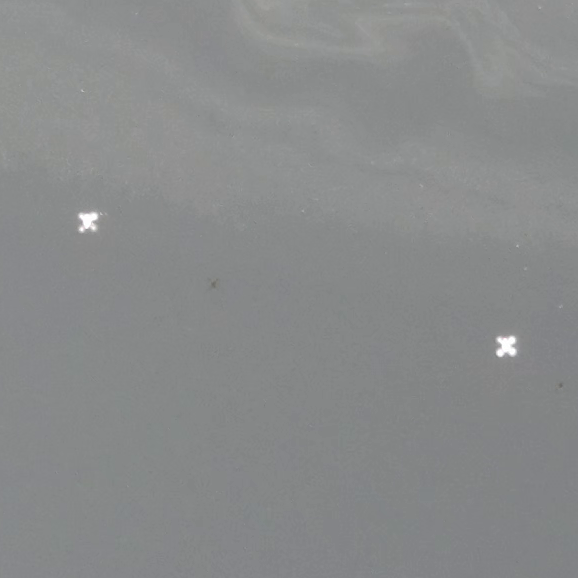}}
   \caption{Snapshots from the real field deployment of the proposed approach.
   Supporting material and media from the experiment are available at~\cite{uavmon}.\label{fig:experiment}}
\end{figure}

The feasibility of the proposed solution has been further examined in an experimental field deployment with limited computational resources.
The proposed approach was experimentally tested with two UAV research platforms based on  DJI F450, shown in~\cref{fig:f450}, performing a water surface exploration task as depicted in~\cref{fig:plan_example}.
The vehicles were equipped with GPS-based navigation and a top-down visual sensor.
The connection between the vehicles and the planning computer was established over a Wi-Fi network using ROS.
The used communication network introduced transport delays due to networking limits over long distances.
Therefore, the parameters of the planning scheme from~\cref{sec:evaluation} were modified to identical re-planning rate and horizon length of $\SI{5}{s}$ and the maximum solver processing time was limited to $t\textsubscript{wall}\leq\SI{3}{s}$. 

After obtaining a solution, time-stamped command trajectories were passed to the UAV trajectory trackers using ROS messages for open-loop monitoring control.
The onboard trackers adjusted to the trajectories on the ROS message arrival.
The planner loop initiated from the UAV states $\SI{5}{s}$ in the future.
The $\SI{2}{s}$ long window left for data transfer ensured the continuity of the flight trajectories.
The vehicles operated over the surface of the Orlík dam, located on the Vltava river in southern Bohemia.
The vehicles are depicted in~\cref{fig:topdown}.
Compared to $r_{\text{eq}} = 18.9$ provided by the ideal planning rate identical to $T_s = 0.25$, the monitoring performance of adjusted real-world setup was $r_{eq}=43.7$ during the $\SI{5}{\min}$ flight.
The performance degradation over an ideal configuration due to the limited computational time and resources can be mitigated using more powerful computational hardware.
Nevertheless, the presented monitoring approach was shown to be feasible in a real-world deployment.

% - section -------------------------------------------------------------------
\section{Conclusion}\label{sec:conclusion}
We propose a novel formulation of the introduced IMP-DR model-based multi-vehicle monitoring approach.
The proposed solution is based on the MPC on receding horizon evaluated on several problem instances.
Based on the reported results, the performance of the proposed approach surpasses the state-of-the-art Kinematic OP solver in static reward collection and the grid-based Team OP solver in dynamic monitoring tasks.
The proposed approach has been utilized in the combined dynamic water surface exploration and monitoring missions with multiple UAVs and limited sensor range.
The results demonstrated the IMP-DR in theory, and the experimental deployment supports its practical viability in monitoring tasks.
Future research is directed at formulating the planning-oriented model prediction to increase performance and address multi-rotor vehicle limited-thrust model and battery charge constraints in prolonged monitoring missions.

\bibliographystyle{IEEEtran}
\bibliography{main}

\end{document}